\documentclass[10pt,twocolumn,letterpaper]{article}

\usepackage[pagenumbers]{cvpr} % 
\usepackage{tikz}
\usetikzlibrary{spy, positioning}
\usepackage{pgfplots}

\usepackage[export]{adjustbox}
\usepackage{amsthm}

 {\everymath{\displaystyle\everymath{}}\array}%
 {\endarray}
\everymath{\displaystyle\everymath{}}

\newtheorem{theorem}{Theorem}[section]
\newtheorem{proposition}[theorem]{Proposition}

\newtheorem*{proposition_non}{Proposition}

\def\A{\mathbf{A}}

\def\I{\mathbf{I}}

\def\e{\mathbf{e}}

\def\g{\mathbf{g}}
\def\h{\mathbf{h}}

\def\w{\mathbf{w}}
\def\x{\mathbf{x}}
\def\y{\mathbf{y}}
\def\z{\mathbf{z}}

\newcommand{\vv}{\mathbf{v}}

\def\bSigma{\boldsymbol{\Sigma}}

\def\bepsilon{\boldsymbol{\epsilon}}

\def\bmu{\boldsymbol{\mu}}

\def\Exp{\mathbb{E}}

\newcommand{\Exppp}[2]{\Exp_{#1}\left[#2\right]}

\def\0{\mathbf{0}}
\def\1{\mathbf{1}}

\newcommand{\tomt}[1]{\textcolor{black}{#1}}
\newcommand{\tomtb}[1]{\textcolor{black}{#1}}

\usepackage{color}
\usepackage{algorithm}
\usepackage{graphicx}
\usepackage{subcaption}
\usepackage{algpseudocode}
\usepackage{wrapfig}
\usepackage{diagbox}
\definecolor{cvprblue}{rgb}{0.21,0.49,0.74}
\usepackage[pagebackref,breaklinks,colorlinks,allcolors=cvprblue]{hyperref}

\def\paperID{15282} % 
\def\confName{CVPR}
\def\confYear{2025}

\title{Zero-Shot Image Restoration Using Few-Step Guidance of Consistency Models (and Beyond)}

\author{Tomer Garber\\
Bar-Ilan University, Israel\\
{\tt\small tomergarber@gmail.com}
\and
Tom Tirer\\
Bar-Ilan University, Israel\\
{\tt\small tirer.tom@gmail.com}
}

\begin{document}
\maketitle

\begin{abstract} 
In recent years, it has become popular to tackle image restoration tasks with a single pretrained diffusion model (DM) and data-fidelity guidance, instead of training a dedicated deep neural network per task. 
However, such ``zero-shot'' restoration schemes currently require many Neural Function Evaluations (NFEs) for performing well, which may be attributed to the many NFEs needed in the original generative functionality of the DMs. 
Recently, faster variants of DMs have been explored for image generation. These include Consistency Models (CMs), which can generate samples via a couple of NFEs.
However, existing works that use guided CMs for restoration still require tens of NFEs or fine-tuning of the model per task 
that leads to performance drop if the assumptions during the fine-tuning are not accurate.
In this paper, we propose a zero-shot restoration scheme that uses CMs 
and operates well with as little as 4 NFEs.
It is based on a wise combination of several ingredients: better initialization, back-projection guidance, and above all a novel noise injection mechanism.
We demonstrate the advantages of our approach for image \tomtb{super-resolution, deblurring} and inpainting.
Interestingly, we show that the usefulness of our noise injection technique goes beyond CMs: it can also mitigate the performance degradation of existing guided DM methods when reducing their NFE count. The source code is available at \href{https://github.com/tirer-lab/CM4IR}{https://github.com/tirer-lab/CM4IR}
\end{abstract}

\begin{figure}[t]
    \centering
    \begin{tikzpicture}[spy using outlines={magnification=2.5, rectangle, size=1cm, blue, connect spies}]
        
        \node (n1) at (-1.3,0) {\includegraphics[width=2.7cm, height=2.7cm]{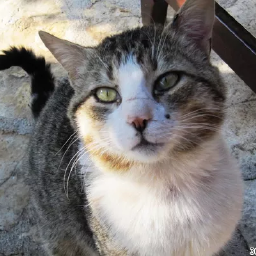}};
        \spy on(-0.9,0.5) in node at (-2.135,-0.835);
        
        \node (n2) at (1.45,0) {\includegraphics[width=2.7cm, height=2.7cm]{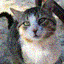}};
        \spy on(1.85,0.5) in node at (0.615,-0.835);
        
        \node (n3) at (4.2,0) {\includegraphics[width=2.7cm, height=2.7cm]{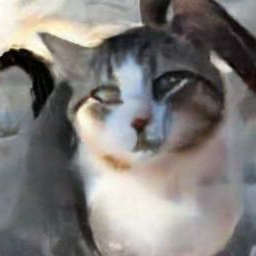}};
        \spy on(4.6,0.5) in node at (3.365,-0.835);
        
        \node (n4) at (-1.3,-2.75) {\includegraphics[width=2.7cm, height=2.7cm]{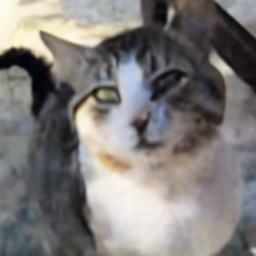}};
        \spy on(-0.9,-2.25) in node at (-2.135,-3.585);
        
        \node (n5) at (1.45,-2.75) {\includegraphics[width=2.7cm, height=2.7cm]{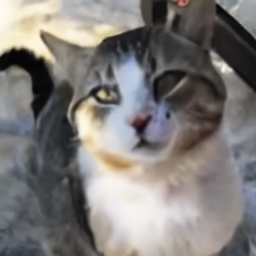}};
        \spy on(1.85,-2.25) in node at (0.615,-3.585);
        
        \node (n6) at (4.2,-2.75) {\includegraphics[width=2.7cm, height=2.7cm]{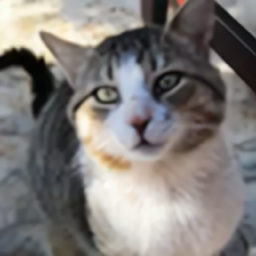}};
        \spy on(4.6,-2.25) in node at (3.365,-3.585);
        
    \end{tikzpicture}

    \caption{
    Super-resolution $\times 4$ with bicubic kernel and noise level of 0.05. From left to right and top to bottom: original, observation,  DPS \cite{chung2022diffusion} (1000 NFEs),
    DiffPIR \cite{zhu2023denoising} (20 NFEs), DDRM \cite{kawar2022denoising} (20 NFEs) and our CM4IR (4 NFEs).}
    \label{fig:cat_sr_0.05_intro}
\end{figure}
\begin{figure}[h]
    \centering
    \begin{tikzpicture}[spy using outlines={magnification=2, rectangle, height=1cm, width=0.7cm, blue, connect spies}]
        
        \node (n1) at (-1.3,0) {\includegraphics[width=2.7cm, height=2.7cm]{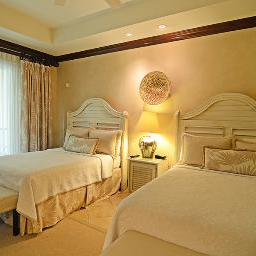}};
        \spy on(-1.08,-0.06) in node at (-2.286,-0.835);
        
        \node (n2) at (1.45,0) {\includegraphics[width=2.7cm, height=2.7cm]{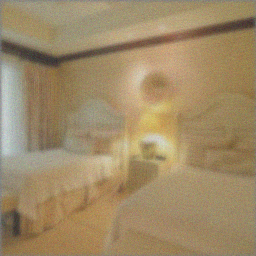}};
        \spy on(1.67,-0.06) in node at (0.465,-0.835);
        
        \node (n3) at (4.2,0) {\includegraphics[width=2.7cm, height=2.7cm]{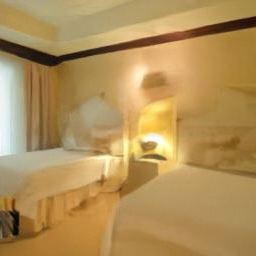}};
        \spy on(4.42,-0.06) in node at (3.215,-0.835);
        
        \node (n4) at (-1.3,-2.75) {\includegraphics[width=2.7cm, height=2.7cm]{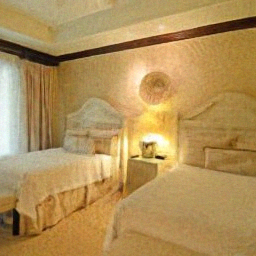}};
        \spy on(-1.08,-2.812) in node at (-2.286,-3.587);
        
        \node (n5) at (1.45,-2.75) {\includegraphics[width=2.7cm, height=2.7cm]{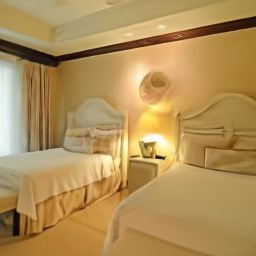}};
        \spy on(1.67,-2.812) in node at (0.465,-3.587);
        
        \node (n6) at (4.2,-2.75) {\includegraphics[width=2.7cm, height=2.7cm]{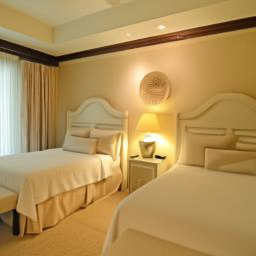}};
        \spy on(4.42,-2.812) in node at (3.215,-3.587);
        
    \end{tikzpicture}
    
    \caption{
    Deblurring with Gaussian kernel and noise level of 0.025. From left to right and top to bottom: original, observation, DPS \cite{chung2022diffusion} (1000 NFEs), 
    DiffPIR \cite{zhu2023denoising} (20 NFEs), DDRM \cite{kawar2022denoising} (20 NFEs) and our CM4IR (4 NFEs).}
    \label{fig:g_deb_0.025_intro}
\end{figure}

\section{Introduction}
\label{sec:intro}

Image restoration problems appear in a wide range of applications, where the goal is to recover a high-quality image $\x^* \in \mathbb{R}^n$ from its degraded version $\y \in \mathbb{R}^m$, which can be incomplete, noisy, blurry, low-resolution, etc. 
In many popular tasks, the relation between $\y$ and $\x^*$ can be expressed using a linear observation model
\begin{align}
\label{eq:obsevation_model}
    \y = \A\x^* + \e
\end{align}
where $\A \in \mathbb{R}^{m\times n}$ is a measurement operator with $m\leq n$, and $\e\sim\mathcal{N}(\0, \sigma_y^2\I_m)$ is an additive white Gaussian noise.
Examples include: denoising, inpainting, deblurring, super-resolution, computed tomography, magnetic resonance imaging, and more. Each of these tasks is associated with a different structure of $\A$.
Importantly, image restoration problems are % 
ill-posed: fitting the observation $\y$ does not ensure a successful recovery and thus prior knowledge on the nature of $\x^*$ needs to be utilized.

Since the breakthroughs in deep learning a decade ago \cite{krizhevsky2012imagenet}, it has become common to address these problems by training a different deep neural network (DNN) for each {\em predefined} observation model in a supervised manner \cite{dong2015image,% 
lim2017enhanced}. Specifically, given a collection of  high-quality images $\{\x^*_i\}$, training pairs $\{\y_i,\x^*_i\}$ are generated by mapping $\x^*_i$ to $\y_i$ using \eqref{eq:obsevation_model}, 
and a DNN is trained to invert the map.
However, these ``task-specific" DNNs % 
suffer from a huge performance drop when the observations at test-time mismatch (even slightly) the assumptions made in training \cite{shocher2018zero,tirer2019super,hussein2020correction,tirer2024deep}, which limits their 
applicability in many practical cases. % 

An increasingly popular alternative to ``task specific'' DNNs are ``zero-shot" approaches, where a pretrained DNN imposes only the signal prior while agreement of the reconstruction with the observations (of potentially arbitrary model) is handled at test-time.
Different choices of such DNNs include plain denoisers \cite{zhang2017beyond} as used in \cite{venkatakrishnan2013plug,zhang2017learning,tirer2018image}, generative adversarial networks \cite{goodfellow2014generative} as used in \cite{bora2017compressed,hussein2020image}, and recently (unconditional) diffusion/score models (DMs) \cite{song2019generative,ho2020denoising} as used in \cite{song2021solving,
kawar2022denoising,song2022pseudoinverse,chung2022diffusion,wang2022zero,abu2022adir,zhu2023denoising,garber2024image}.

Despite their flexibility to the observation model, the main limitation of zero-shot restoration schemes that are based on  DMs is that they require many Neural Function Evaluations (NFEs) of highly overparameterized models. % 
The core reason for this is inherited from the original generative functionality of the DMs, which itself requires many NFEs. 
Indeed, the restoration schemes are based on equipping sampling schemes, consisting of many iterations of denoising and noise injection, with data-fidelity guidance. 
For a given DM, sophisticated guidance techniques can reduces the number of NFEs from 1000 NFEs \cite{chung2022diffusion} to a couple of dozens \cite{kawar2022denoising,zhu2023denoising}, % 
but to our knowledge not less than this.
More details on the connections between types of data-fidelity guidance, their acceleration aspect, and the relation of guided DMs to proximal  algorithms can be found in \cite{garber2024image}.

Attempts to reduce the number of NFEs required for image generation is an active research topic.
Some works focus on modifying the sampling schemes (``reversed flows'') of pretrained DM \cite{song2020denoising,karras2022elucidating}. A greater acceleration is achieved using new training strategies for the denoising models \cite{salimans2022progressive,abu2023udpm,song2023consistency}.
Among those, a prominent alternative to DMs are Consistency Models (CMs) \cite{song2023consistency}, which can generate samples via a couple of NFEs. 
However, existing works that use guided CMs for restoration still require tens of NFEs \cite{song2023consistency} or fine-tuning of the model per task \cite{zhao2024cosign}. 
Note that the latter is not a zero-shot strategy
and, as such, \cite{zhao2024cosign} is prone to suffer from significant performance drop if the assumptions
during the fine-tuning (e.g., the noise level) are not accurate.

\textbf{Contribution.} 
In this paper, we propose a zero-shot restoration scheme that utilizes CMs and, with as little as 4 NFEs, outperforms alternative zero-shot methods (which use many more NFEs) on popular tasks such as \tomtb{image super-resolution, deblurring and inpainting}.
Our approach is based on a wise combination of several ingredients: better initialization, back-projection guidance, and above all a novel noise injection mechanism.

Our noise injection mechanism deviates from existing techniques in two aspects that allow us to reduce the number of iterations. First, we  decouple the noise level of the denoising operation from the noise level of the injection, such that it takes into account noise that originates from the data-fidelity guidance and gives the denoiser more ``freedom'' to modify its input.
Second, we split the amount of injected noise between the stochastic noise and an estimated noise ({\em anti}-correlated to the one in \citep{song2020denoising}) that boosts the sampling scheme.
We formally motivate this split by: 1) showing that it preserves  properties of marginal distribution of the samples when omitting the guidance (the aforementioned noise decoupling takes the guidance into account), and 2) identifying it as a ``noisy variant'' of Polyak acceleration \cite{polyak1964some}.

Finally, we show that the usefulness of our noise injection technique {\em goes beyond CMs}: it can also mitigate the performance degradation of existing guided DM methods \cite{kawar2022denoising,zhu2023denoising} when the number of iterations (NFEs) that they
use is drastically reduced.

\section{Background and Related Work}
\label{sec:background}

\subsection{DMs and CMs}
\label{sec:background_DMs}

Let $p_{\text{data}}(\x)$ denote the data distribution.
Diffusion/score-based generative models \cite{song2019generative,ho2020denoising,song2020score,karras2022elucidating} 
are based on reversing a user-designed ``forward flow'' 
where $p_{\text{data}}$ is diffused till it reaches the tractable Gaussian distribution $\mathcal{N}(\0,\sigma_{\text{max}}^2\I)$. The forward flow can be expressed by the stochastic differential equation (SDE)
\begin{align}
\label{eq:forward_sde}
    d\x_t = \h(\x_t,t)\mathrm{d}t + g(t) \mathrm{d} \w_t,
\end{align}
where $t \in [0,T]$ with constant $T>0$, $\h$ and
$g$ are the drift and diffusion coefficients, respectively, and $\{ \w_t \}$ denotes the standard Brownian motion.
We denote the distribution of $\x_t$ by $p_t(\x)$, so $p_0=p_{\text{data}}$ and we aim for $p_T = \mathcal{N}(\0,\sigma_{\text{max}}^2\I)$.

As pointed out by \cite{song2020score}, based on a result of \cite{anderson1982reverse}, the forward SDE is associated with a reversed SDE 
\begin{align}
\label{eq:reversed_sde}
    d\x_t = \left[ \h(\x_t,t) - g(t)^2 \nabla_{\x}\log p_t(\x_t) \right ] \mathrm{d}t + g(t) \mathrm{d} \tilde{\w}_t,
\end{align} 
where $\tilde{\w}$ is a standard Brownian motion when time flows backwards from $T$ to 0, and $\mathrm{d}t$ is an infinitesimal negative timestep. 
Interestingly, \cite{song2020score} showed that the reversed SDE has an associated ordinary differential equation (ODE), dubbed Probability Flow (PF), whose solution trajectories sampled at $t$ are distributed according to $p_t(\x)$:
\begin{align}
\label{eq:reversed_PF}
    d\x_t = \left[ \h(\x_t,t) - \frac{1}{2}g(t)^2 \nabla_{\x}\log p_t(\x_t) \right ] \mathrm{d}t.
\end{align} 
One can approximately obtain a sample of $p_{\text{data}}$ from an initial $\x_T \sim p_T$ in a ``reversed flow'', provided that he has an accurate estimator of $\nabla_{\x}\log p_t(\cdot)$, known as the \textit{score function} of $p_t$, and sufficiently good discretization and integration of the reversed SDE/ODE.

In DMs, a common choice is $\h(\x_t,t)=\0$, known as the VE-SDE formulation \cite{song2020score,karras2022elucidating}. In this case, by \eqref{eq:forward_sde}, $\x_t$ is a noisy version of an unscaled data sample $\x_0 \sim p_0$ with additive Gaussian noise of some level $\sigma_t$, depending on the choice of $g(t)$. 
For $\x_t = \x_0 + \sigma_t\bepsilon_t$ with $\bepsilon_t \sim \mathcal{N}(\0,\I)$,
we have that $-\sigma_t^2 \nabla_{\x}\log p_t(\x_t) = \x_t - \mathbb{E}[\x_0|\x_t] =\sigma_t\mathbb{E}[\bepsilon_t|\x_t]$,
where 
the first equality is Tweedie's formula \cite{efron2011tweedie} and the second equality follows from $\mathbb{E}[\x_0|\x_t]=\mathbb{E}[\x_t - \sigma_t\bepsilon_t|\x_t] = \x_t - \sigma_t \mathbb{E}[\bepsilon_t|\x_t]$. 
Therefore, conceptually, there is equivalence between learning: % 
1) the score function \cite{song2019generative,song2020score}, 2) the MMSE denoiser \cite{karras2022elucidating}, or 3) the MMSE of the noise \cite{ho2020denoising}.
Similar equivalence, up to scaling factors, holds in the VP-SDE formulation, where $\h(\x_t,t)=-\frac{1}{2}\beta(t)\x_t$ and $g(t)=\sqrt{\beta(t)}$ \cite{ho2020denoising,song2020score}. 

A denoiser $f_{\theta}( \x, \sigma)$, where $\x$ is the signal and $\sigma$ is the noise level, can be parameterized in different ways. 
For example, \cite{karras2022elucidating} parameterized it as 
\begin{align}
\label{eq:D_formula}
f_{\theta}( \x, \sigma) = c_{\text{skip}}(\sigma)\x + c_{\text{out}}(\sigma)F_{\theta} \left (c_{\text{in}}(\sigma)\x,c_{\text{noise}}(\sigma)\sigma \right ),    
\end{align}
where $F_{\theta}$ is the DNN and $c_{\text{skip}}(\cdot), c_{\text{out}}(\cdot), c_{\text{in}}(\cdot), c_{\text{noise}}(\cdot)$ are coefficient functions that are set manually.
Their denoiser is trained for the range of noise level $\sigma$ that is required for the reversed flow by minimization of
\begin{align}
\label{eq:D_opt}
    \Exppp{\sigma,\x_0\sim p_{\text{data}},\x \sim \mathcal{N}(\x_0,\sigma^2\I)}{\lambda(\sigma) d \left ( f_{\theta}( \x, \sigma) , \x_0 \right )}
\end{align}
where $\lambda(\sigma)$ is a weighting function the loss and the distance metric is  $d(\x,\tilde{\x})=\|\x-\tilde{\x}\|_2^2$, aligned with MMSE estimation.

After training, the sampling schemes are based on numerical solvers (discretization and integration) of the reversed SDE \eqref{eq:reversed_sde}, % 
which boil down
to random Gaussian initialization of $\x_T$ and iterations of denoising and noise injection with decreasing noise levels. 
Many works also use the choice $g(t)=\sqrt{2t}$, which leads to a simple $\sigma_t=t$ \cite{song2020score,karras2022elucidating,song2023consistency}. 
Then, for a sequence of time points $T=\tau_N>\tau_{N-1}>\ldots>\tau_2>\tau_1=\epsilon$ (discretization of $[\epsilon,T]$ with $\epsilon \approx 0_+$ and typically $T=1$) and initialization $\x_{\tau_N} \sim \mathcal{N}(\0,\tau_N^2\I)$, a typical sampling scheme with $n=N,N-1,...,1$ reads as
\begin{align}
\label{eq:sampling_est_x0}
    &\x_{0|\tau_{n}} = f_{\theta}( \x_{\tau_n}, \tau_n) \\
\label{eq:sampling_noise}
    &\z \sim \mathcal{N}(\0,\I) \\
\label{eq:sampling_update}    
    &\x_{\tau_{n-1}} = \x_{0|\tau_{n}} + \tau_{n-1} \z
\end{align}

Attempts to reduce the overall number of NFEs in the sampling scheme include improving the estimated denoised signal using two NFEs per iteration \cite{karras2022elucidating} (i.e., second-order integrator) or replacing the noise injection in \eqref{eq:sampling_update} with the DDIM procedure \cite{song2020denoising}:
\begin{align}
\label{eq:noise_inj_ddim_style}
    \sqrt{1-\eta^2} \tau_{n-1} \hat{\z} + \eta \tau_{n-1} \z,
\end{align}
where $\eta \in [0,1]$ is a hyperparameter that trades between the stochastic noise $\z$ and the estimated noise
\begin{align}
\label{eq:noise_estimate}
    \hat{\z} = ( \x_{\tau_{n}} - f_{\theta}( \x_{\tau_n}, \tau_n))/\tau_{n}.
\end{align}
Note \eqref{eq:noise_inj_ddim_style} slightly differs from the presentation in \cite{song2020denoising}, since they use VP-SDE formulation and a noise estimator model (i.e., ${\z}_{\theta}( \x_{\tau_n}, \tau_n)$ rather than $\hat{\z}$ and $f_{\theta}( \x_{\tau_n}, \tau_n)$).

However, despite these efforts, 
with a denoiser (or noise estimator) optimized as in \eqref{eq:D_opt}, 
even techniques that use more sophisticated updates require not less than three dozens NFEs for high-quality image generation \cite{karras2022elucidating}.

To achieve a significant decrease in the number of NFEs, recent works have proposed different training strategies for the denoising models \cite{salimans2022progressive,abu2023udpm,song2023consistency}.
Among them, Consistency Models (CMs) \cite{song2023consistency} has received much attention.

CMs attempt to train a denoiser $f_{\theta}$ in a way that better promotes that it outputs the same clean image $\x_0$ for different points along path of the PF ODE \eqref{eq:reversed_PF} from $\x_T$ to $\x_0$. % 
To this end, the most performant training technique in \cite{song2023consistency} is based on distilling a pre-trained DM $\Phi$ to obtain by deterministic numerical integration $\hat{\x}^{\Phi}_{t_n}$, an approximation of the point $\x_{t_n}$ on the PF path from $\x_T$ to $\x_0$, given $\x_0$. 
For a predetermined sequence for training $T=t_N>t_{N-1}>\ldots>t_2>t_1=\epsilon$, which may differ from the one used for sampling, the common training loss \eqref{eq:D_opt} (with $\sigma=t$) is replaced with 
\begin{align}
\label{eq:CD_loss}
    &\mathbb{E}_{n\sim U[1,N-1],\x_0\sim p_{\text{data}},\x_{t_{n+1}} \sim \mathcal{N}(\x_0,t_{n+1}^2\I)} \\ \nonumber
    &\hspace{10mm} \left [ \lambda(t) d \left ( f_{\theta}( \x_{t_{n+1}}, t_{n+1}) , f_{\theta^-}( \hat{\x}^{\Phi}_{t_n}, t_{n}) \right ) \right ],
\end{align}
where $f_{\theta^-}$ is a ``target'' copy of the denoiser whose parameters $\theta^-$ are exponential moving average of $\theta$ with $\mathrm{stopgrad}$.
To avoid degenerate constant output of $f_{\theta}$, a specific parameterization \eqref{eq:D_formula} is used: $c_{\text{skip}}(\epsilon)=1$ and $c_{\text{out}}(\epsilon)=0$, which ensures that $f_{\theta}( \x, \epsilon)=\x$.
We refer the reader to \cite{song2023consistency} for additional details on the architecture, coefficients functions and the training procedure of CMs. We note that \cite{song2023consistency} proposes also an approach for training the denoiser without distillation, but the aforementioned approach performs better, so we choose to consider CMs trained in this way.

Remarkably, CMs have been shown to generate quite good images already with 1 NFE, i.e., simply by $\x_T \sim \mathcal{N}(\0,T^2\I)$ and $\x_{0|T} = f_{\theta}( \x_T, T)$.
The results are improved when using the iterative sampling \eqref{eq:sampling_est_x0}-\eqref{eq:sampling_update} even with very small $N$ and well chosen $\{ \tau_n \}$.

\subsection{Restoration via Guidance of DMs}

Pretrained DMs have been shown to be a powerful signal prior for image restoration \cite{song2021solving,kawar2022denoising,song2022pseudoinverse,chung2022diffusion,wang2022zero,abu2022adir,zhu2023denoising,garber2024image}. 
In order that the sampling scheme will produce an image that is not only perceptually pleasing but also agrees with the measurements $\y$ and the observation model \eqref{eq:obsevation_model}, it is required to equip the iterations with some data-fidelity guidance.
This guidance is typically based on the gradient of a data-fidelity term $\ell(\x;\y)$. In other words, the update \eqref{eq:sampling_update} (oftentimes with the DDIM noise injection \eqref{eq:noise_inj_ddim_style} \citep{kawar2022denoising,wang2022zero,zhu2023denoising,garber2024image})
is modified into 
\begin{equation}
\label{eq:sampling_update_restoration}
    \x_{\tau_{n-1}} = \x_{0|\tau_{n}} - \mu_n \nabla_{\x} \ell( \x_{0|\tau_{n}} ;\y) + \tau_{n-1} \z,
\end{equation}
where $\mu_{n}$ is the guidance scaling factor. % 

Least squares (LS) guidance is a typical choice
\begin{align}
\label{eq:ls_term}
    &\ell_{\text{LS}}(\x;\y) = \frac{1}{2}\|\A\x-\y\|_2^2, \\
    &\nabla_{\x} \ell_{\text{LS}}(\x;\y) = \A^T(\A\x - \y),
\end{align}
which is used, e.g., in % 
\cite{chung2022diffusion}.
Yet, acceleration (less NFEs) is obtained with the back-projection (BP) guidance \cite{tirer2018image,tirer2019back}
\begin{align}
\label{eq:bp_term}
    &\ell_{\text{BP}}(\x;\y) = \frac{1}{2}\|(\A\A^T)^{-1/2}(\A\x-\y)\|_2^2, \\
\label{eq:bp_step}    
    &\nabla_{\x} \ell_{\text{BP}}(\x;\y) = \A^{\dagger}(\A\x - \y),
\end{align}
where $\A^\dagger=\A^T(\A\A^T)^{-1}$ is the pseudoinverse of $\A$, which can be regularized and oftentimes has efficient implementation (generally by conjugate gradients \cite{hestenes1952methods}, and in super-resolution and deblurring even by FFT \cite{garber2024image}). 
See \cite{tirer2021convergence} for theoretical analysis of the acceleration property. 
This guidance has been used recently in \cite{song2021solving,song2022pseudoinverse,wang2022zero,garber2024image}, sometimes under different names (e.g., ``pseudoinverse guidance'').
When the SVD of $\A$ can be computed efficiently, a guidance that resembles BP but can mitigate noise amplification per singular component has been used in the DDRM method \cite{kawar2022denoising}.
Importantly, existing zero-shot restoration techniques based on guided DMs still require at least a couple of dozens NFEs.

CMs pave the way to zero-shot restoration techniques with less NFEs.
However, existing works that use guided CMs \cite{song2023consistency,zhao2024cosign} still have limitations. Specifically, the guided scheme in the original CM paper \cite{song2023consistency} requires 40 NFEs.
In the very recent CoSIGN paper \cite{zhao2024cosign}, the authors apply supervised fine-tuning to the model per task, which allows them to restore perceptually pleasing images with a couple of iterations via BP-guidance. Yet, each NFE in \cite{zhao2024cosign} is more computationally heavy as a ControlNet \cite{zhang2023adding} is integrated to the original pretrained CM.
Furthermore, and more crucially, this is not a zero-shot strategy and, as such, leads to performance drop if the assumptions during the fine-tuning (e.g., the noise level) mismatch the situation at test-time.

\section{The Proposed Method}
\label{sec:method}

In this paper, we propose a few-step zero-shot restoration technique that utilizes pretrained CMs.
Our approach combines several ingredients: initialization, BP guidance, and above all a 
novel noise injection mechanism, each is described below.

\textbf{Initialization.} 
Most of the DM based restoration techniques initialize $\x_{\tau_N}$ with pure noise $\x_{\tau_N} \sim \mathcal{N}(\0,T^2\I)$, or equivalently
\begin{align}
    \x_{\tau_N} = \x_{\text{init}} + \tau_N\z
\end{align}
with $\z \sim \mathcal{N}(\0,\I)$, $\x_{\text{init}}=\0$ and $\tau_N=T$.
This ignores the fact that the observations vector $\y$ contains information on the specific $\x^*$ that we wish to restore. 
Therefore, in general, we propose to set $\x_{\text{init}}=\A^{\dagger}\y$.
For example, if $\A$ is bicubic downsampling (in super-resolution task), then $\A^{\dagger}$ is bicubic upsampling.
For inpainting, though, $\A$ is subsampled rows of the identity matrix and $\A^{\dagger}=\A^T$ just fills the missing pixels with zeros, so we propose to use a better choice: median initialization as in \cite{tirer2018image}.
Our initialization resembles to the common practice in many methods prior to DMs \cite{venkatakrishnan2013plug,zhang2017learning,tirer2018image}. The difference from these works, though, is that we still add noise of level $\tau_N$ to $\x_{\text{init}}$. To avoid masking $\x_{\text{init}}$ at initialization, the noise level $\tau_N$ will be smaller than $T$.

\textbf{Guidance.}
Following the acceleration property of back-projections  \cite{tirer2018image,tirer2019back,tirer2021convergence}, we choose to use the BP guidance \eqref{eq:bp_step}. 
Note that at high noise levels in $\y$ this may require using decaying step sizes $\{ \mu_n \}$.

\textbf{Noise injection.}
We propose a novel noise injection mechanism that deviates from existing techniques in two aspects.

First, we claim that in restoration tasks, the noise level of the denoising operation and the noise level of the injection do not need to be the same.
This is due to: 1) If $\y$ contains noise, $\sigma_y>0$, then the guidance $\nabla_{\x} \ell( \x_{0|\tau_{n}} ;\y)$ adds to the sample $\x_{\tau_{n-1}}$ a noise component that originates from $\y$; and 2) Regardless of $\sigma_y$, at early iterations the estimated signal can differ significantly from the underlying signal $\x^*$, and thus by increasing the denoiser's noise level above the value used for the injection noise, we give the denoiser more ``freedom'' in modifying its input, which in return allows us to reduce the number of iterations. 
{Therefore, for noise injection of level $\tau_n$ we propose to apply the CM $f_{\theta}$ with noise level \tomt{$(1+\delta)\tau_n$}, where $\delta \geq 0$ is a hyperparameter.}

Second, to further accelerate the restoration with CMs, ideally we would like to push the sample $\x_{\tau_{n}}$ towards $\x_{0}$, i.e., in the direction $\x_{0} - \x_{\tau_{n}}$, beyond what we get from one denoising operation of CM.
However, we do not have $\x_{0}$ but rather the denoiser's output $\x_{0|\tau_n}$. 
This motivates us to define:
\begin{align}
    \hat{\z}^{-}:=(\x_{0|\tau_n} - \x_{\tau_{n}})/\tau_{n}.
\end{align}
For $\x_t = \x_0 + t\z$ with $\z \sim \mathcal{N}(\0,\I)$, we have that $\mathbb{E}[\z|\x_t]=\mathbb{E}[(\x_t-\x_0)/t|\x_t]=(\x_t - \mathbb{E}[\x_0|\x_t])/t$. Considering $\x_{0|t}$ as a replacement of $\mathbb{E}[\x_0|\x_t]$, we can understand $\hat{\z}^-$ as the \textit{negative} estimate of the noise.
Note that the sign does not affect the similarity of $\hat{\z}^-$ to Gaussian noise. 
Thus, we propose to split the noise injection between the stochastic noise $\z \sim \mathcal{N}(\0,\I)$ and the estimated $\hat{\z}^-$. 
The proposed noise injection for generating $\x_{\tau_{n-1}}$ reads as
\begin{align}
\label{eq:noise_inj}
    \sqrt{1-\eta^2} \tau_{n-1} \hat{\z}^{-} + \eta \tau_{n-1} \z
\end{align}
where $\eta \in [0,1]$ is a hyperparameter that trades between the two types of noise.

\textbf{The full algorithm.} 
The proposed technique, CM for Image Restoration (CM4IR), is summarized in Algorithm~\ref{alg:CM_bp}.
In Section \ref{sec:exp} we will show its performance with a choice of $N=4$ values of $\{\tau_n\}$, i.e., with 4 NFEs.

\textbf{Relation of \eqref{eq:noise_inj} to DDIM.}
As shown in \eqref{eq:noise_inj_ddim_style}, recall that
the DDIM scheme \cite{song2020denoising} accelerates a baseline scheme \cite{ho2020denoising} % 
by noise injection similar to \eqref{eq:noise_inj} but with $\hat{\z}=-\hat{\z}^{-}$ instead of $\hat{\z}^{-}$ (i.e., anti-correlation).
In \cite{song2020denoising},  
$\hat{\z}$ is motivated as being a ``direction pointing to the current $\x_{\tau_{n}}$''. 
We claim that % 
\textcolor{black}{for few-step restoration,} 
attracting the update back to the current $\x_{\tau_{n}}$, by using $\hat{\z}$ instead of $\hat{\z}^{-}$ is not beneficial. This will be demonstrated empirically in Section \ref{sec:exp_ablation}.
All the more so, in Section \ref{sec:exp_DMs} we will show that this simple modification mitigates the performance drop of guided DM methods \cite{kawar2022denoising,zhu2023denoising}, which are based on DDIM scheme, when drastically reducing the number of the NFEs.

Furthermore, in \cite{song2020denoising}, a major motivation for the procedure \eqref{eq:noise_inj_ddim_style} (with $\eta < 1$) is that % 
under their assumption on the distribution $q_{\eta}(\x_{\tau_1:\tau_N}|\x_0)$, for any $n>1$
the  distribution\footnote{We modified the notations of \cite{song2020denoising} from VP-SDE to VE-SDE.} 
\begin{align}
\label{eq:ddim_distribution}
    &q_{\eta}(\x_{\tau_{n-1}}|\x_{\tau_{n}},\x_{0}) \nonumber \\ 
    &= \mathcal{N}\left( \x_0 + \sqrt{1-\eta^2}\tau_{n-1} \frac{\x_{\tau_n} - \x_0}{\tau_n} , \eta^2\tau_{n-1}^2\I \right)
\end{align}
from which their reversed flow is derived (by replacing $\x_0$ with its current estimate)
possesses the ``desired'' marginal
$q_{\eta}(\x_{\tau_{n-1}}|\x_{0}) = \mathcal{N}(\x_0,\tau_{n-1}^2\I)$, which matches pretrained DMs (no dependency on $\eta$).
This property is preserved in our case, as stated below and proved in the appendix.

\begin{proposition}
\label{prop:marginal}
    Under the assumption of \cite{song2020denoising} on $q_{\eta}(\x_{\tau_1:\tau_N}|\x_0)$, we have $q_{\eta}(\x_{\tau_{n-1}}|\x_{0}) = \mathcal{N}(\x_0,\tau_{n-1}^2\I)$ also if we replace $(\x_{\tau_n} - \x_0)/\tau_n$ with $(\x_0-\x_{\tau_n})/\tau_n$ in \eqref{eq:ddim_distribution}.
\end{proposition}

\begin{algorithm}[t]
    \caption{CM for Image Restoration (CM4IR)}
    \label{alg:CM_bp}
    \begin{algorithmic}[1]
        \Require $f_\theta(\cdot,t)$ (CM denoiser), $N$, $\{ \tau_n \}$, $\{\mu_n\}$, $\delta$, $\eta$, $\A$, $\y$.
        
        \State
        Initialize $\x_{\tau_N} \sim \mathcal{N}(\A^{\dagger}\y, \tau_N^2\I_n)$ % 

        \For {$n$ from $N$ to $1$}
            \State 
            $\x_{0|\tau_{n}} = f_{\theta}( \x_{\tau_n}, (1+\delta)\tau_n)$

            \State
            $\g_{\text{BP}} = \A^{\dagger}(\A\x_{0|t} - \y)$
            
            \State
            $\hat{\z}^{-} =(\x_{0|\tau_n} - \x_{\tau_{n}})/\tau_{n}$
            
            \State
            $\z \sim \mathcal{N}(\mathbf{0}, \I_n)$

            \State
            $\x_{\tau_{n-1}} = \x_{0|\tau_{n}} - \mu_n \g_{\text{BP}} + \sqrt{1-\eta^2} \tau_{n-1} \hat{\z}^{-} + \eta \tau_{n-1} \z$
            
        \EndFor \\
        \Return $\x_{0|\tau_{1}}$
    \end{algorithmic}
\end{algorithm}

\textbf{Relation to Polyak acceleration.}

In his seminal work \cite{polyak1964some}, 
Polyak suggested to accelerate the gradient descent optimization of an objective, e.g., $\ell(\x)$, by the iterates (with a  decreasing sequence of indices $\tau_N,\tau_{N-1},...,\tau_1$):
\begin{align}
\label{eq:polyak}
\x_{\tau_{n-1}} = \x_{\tau_{n}} - \mu \nabla \ell(\x_{\tau_n}) + \beta (\x_{\tau_n} - \x_{\tau_{n+1}})
\end{align}
where $\mu$ and $\beta$ are step-size and momentum hyperparameters, respectively. 
With suitable choice of hyperparameters, faster convergence than plain gradient descent can be proved for quadratic objective functions \cite{polyak1964some,polyak1987introduction}.  Empirically, the approach can be beneficial in other settings.

When using a prior, e.g., a denoiser $f_\theta(\cdot,t)$, a projected/proximal version of \eqref{eq:polyak} reads as:
\begin{align}
&\x_{0|\tau_n} = f_\theta(\x_{\tau_n},\tau_n) \\
\label{eq:polyak_prox}
&\x_{\tau_{n-1}} = \x_{0|\tau_n} - \mu \nabla \ell(\x_{0|\tau_n}) + \beta (\x_{0|\tau_n} - \x_{0|\tau_{n+1}})
\end{align}

Comparing \eqref{eq:polyak_prox} with Algorithm \ref{alg:CM_bp} (line 7), we see that apart from the stochastic noise $\z$, the main difference is that our $\hat{\z}^{-} \propto \x_{0|\tau_n} - \x_{\tau_{n}}$ while the direction of Polyak acceleration is $\vv := \x_{0|\tau_n} - \x_{0|\tau_{n+1}}$. 
Notice that $\x_{\tau_{n}}$ can be understood as a noisy version of $\x_{0|\tau_{n+1}}$ --- and thus $\hat{\z}^{-}$ is a noisy version of $\vv$.
We claim that such a noise component is essential for our restoration scheme. 
\tomtb{Indeed, with properly scaled noise,  the input to the denoiser in the next iteration better matches the data that the denoiser has been trained on.}  
This will be demonstrated empirically in Section \ref{sec:exp_ablation} by replacing $\hat{\z}^{-}$ with $\beta\vv$ in Algorithm \ref{alg:CM_bp} (line 7) and tuning $\beta$ in addition to other hyperparameters.

\begin{table*}
\scriptsize % 
\renewcommand{\arraystretch}{1.3}
\caption{Ablation study on super-resolution with 4 NFEs.  PSNR [dB] ($\uparrow$) and LPIPS ($\downarrow$) results on LSUN Bedroom validation set.} 
\label{table:results_and_ablation_lsun_bedroom}
\centering
    \begin{tabular}{ | c ||
    c | c | c | c  | c | }
    \hline
 \diagbox[height=2em,width=10em]{Task}{Method} 
 & Alg.1 with $\delta$=0 and $\eta$=1 
  & Alg.1 with $\delta$=0 and $\hat{\z}$ i/o $\hat{\z}^-$   
  & Alg.1 with $\delta$=0 and  $\beta\vv$ (Polyak) i/o $\hat{\z}^-$ % 
  & Alg.1 with $\delta$=0
  & CM4IR 

 \\ \hline \hline

    SRx4~$\sigma_y$=0.025 
    & 24.49 / 0.349
    & 24.64 / 0.348    
    & 23.37 / 0.367
    & 25.94 / 0.298
    & {\bf 26.14} / {\bf 0.295}
     \\ \hline

    SRx4~$\sigma_y$=0.05
    & 23.37 / 0.361
    & 20.95 / 0.606    
    & 22.30 / 0.434
    & 25.51 / {\bf 0.320}
    & {\bf 25.60} / {\bf 0.320}
     \\ \hline

    \end{tabular}

\vspace{5mm}

\scriptsize % 
\renewcommand{\arraystretch}{1.3}
\caption{Super-resolution, deblurring and inpainting. PSNR [dB] ($\uparrow$) and LPIPS ($\downarrow$) results on LSUN Bedroom validation set.} 
\label{table:results_lsun_bedroom}
\centering
    \begin{tabular}{ | c || c | c | c | c  | c|c|c|
    }
    \hline
 \diagbox[height=2em,width=10em]{Task}{Method} & CM (40 NFEs)  % 
 & CoSIGN (task spec.) & DDRM (20 NFEs) % 
 & DiffPIR (20 NFEs) 
 & CM4IR (Ours, 4 NFEs) 
 \\ \hline \hline

    SRx4~$\sigma_y$=0.025 
    & 24.66 / 0.344 
    & 26.10 / {\bf 0.205}
    & 25.67 / 0.316
    & 25.09 / 0.374
    & {\bf 26.14} / 0.295
    \\ \hline

    SRx4~$\sigma_y$=0.05 &  23.62 / 0.449
    & 20.35 / 0.569
    & 25.08 / 0.354
    & 23.83 / 0.457
    & {\bf 25.60} / {\bf 0.320}
    \\ \hline

    Gauss. Deblurring~$\sigma_y$=0.025 
    &  26.07 / 0.339
    &  19.74 / 0.342
    & 28.94 / 0.221
    & 27.48 / 0.319
    & {\bf 29.00} / {\bf 0.213}
    \\ \hline

    Gauss. Deblurring~$\sigma_y$=0.05
    &  24.18 / 0.453
    &  19.08 / 0.543
    & 27.35 /  0.280
    & 26.14 / 0.363
    & {\bf 27.42} / {\bf 0.258}
    \\ \hline

    Inpaint. (80\%) $\sigma_y$=0 \
    &   22.39 / 0.366
    & 23.16 / 0.397
    & 19.40 / 0.545
    & 22.78 / 0.464
    & {\bf 25.43} / {\bf 0.284}
    \\ \hline
    
    Inpaint. (80\%) $\sigma_y$=0.025 
    &  22.17 / 0.417
    & 23.22 / 0.368
    & 19.16 / 0.548
    & 22.65 / 0.477
    & {\bf 25.34} / {\bf 0.295}
    \\ \hline

     Inpaint. (80\%) $\sigma_y$=0.05
     &  21.56 / 0.476
    & 23.22 / 0.442
    & 19.09 / 0.560
    & 22.38 / 0.496
    & {\bf 25.28} / {\bf 0.328}
    \\ \hline
    \end{tabular}

\vspace{5mm}

\scriptsize % 
\renewcommand{\arraystretch}{1.3}
\caption{Super-resolution, deblurring and inpainting. PSNR [dB] ($\uparrow$) and LPIPS ($\downarrow$) results on LSUN Cat validation set.} 
\label{table:results_lsun_cat}
\centering
    \begin{tabular}{ | c || c | c | c | c  | c|c|c|
    }
    \hline
 \diagbox[height=2em,width=10em]{Task}{Method} & CM (40 NFEs)  % 
 & CoSIGN (task spec.) & DDRM (20 NFEs) % 
 & DiffPIR (20 NFEs) 
 &
 CM4IR (Ours, 4 NFEs)
 \\ \hline \hline

    SRx4~$\sigma_y$=0.025
    & 25.63 / 0.366
    & N/A
    & 26.93 / 0.329
    & 26.70 / 0.349 
    & {\bf 27.18} / {\bf 0.328}
    \\ \hline

    SRx4~$\sigma_y$=0.05 
    &  24.03 / 0.459
    & N/A
    & 26.05 / 0.371
    & 25.45 / 0.399
    & {\bf 26.53} / {\bf 0.349}
    \\ \hline

    Gauss. Deblurring~$\sigma_y$=0.025 
    &  26.69 / 0.346
    &  N/A
    & {\bf 29.84} / 0.258
    & 27.93 / 0.330
    & 29.76 / {\bf 0.247} % 
    \\ \hline

    Gauss. Deblurring~$\sigma_y$=0.05
    &  24.54 / 0.453
    &  N/A
    & {\bf 28.33} / 0.316
    & 26.64 / 0.370
    & 28.06 / {\bf 0.299} % 
    \\ \hline

    Inpaint. (80\%) $\sigma_y$=0.025
    & 21.89 / 0.478
    & N/A   
    & 18.51 / 0.648
    & 22.78 / 0.498
    & {\bf 25.89} / {\bf 0.364}
    \\ \hline
    
    Inpaint. (80\%)        $\sigma_y$=0.05 
    & 21.07 / 0.523
    & N/A
    & 18.48 / 0.649
    &  22.49 / 0.514
    & {\bf 25.34} / {\bf 0.423}
    \\ \hline
    \end{tabular}
\end{table*}

\begin{figure*}
    \centering
    \begin{subfigure}[h]{0.138\textwidth}
        \centering
        \includegraphics[width=2.4cm, height=2.4cm]{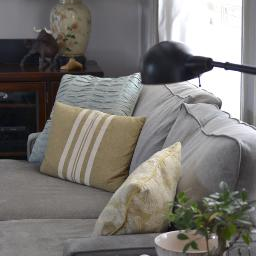}

    \end{subfigure} % 
    \begin{subfigure}[h]{0.138\textwidth}
        \centering
        \includegraphics[width=2.4cm, height=2.4cm]{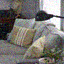}
    \end{subfigure}
    \begin{subfigure}[h]{0.138\textwidth}
        \centering
        \includegraphics[width=2.4cm, height=2.4cm]{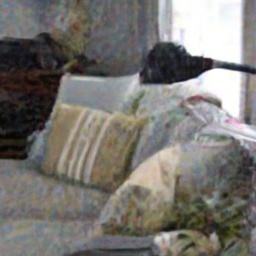}
    \end{subfigure}     
    \begin{subfigure}[h]{0.138\textwidth}
        \centering
        \includegraphics[width=2.4cm, height=2.4cm]{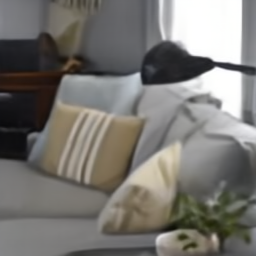}
    \end{subfigure}   
    \begin{subfigure}[h]{0.138\textwidth}
        \centering
        \includegraphics[width=2.4cm, height=2.4cm]{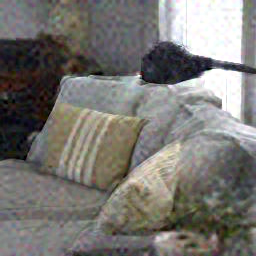}
    \end{subfigure}     
    \begin{subfigure}[h]{0.138\textwidth}
        \centering
        \includegraphics[width=2.4cm, height=2.4cm]{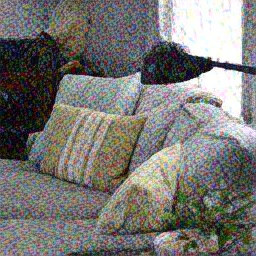}
    \end{subfigure}   
    \begin{subfigure}[h]{0.138\textwidth} 
        \centering
        \includegraphics[width=2.4cm, height=2.4cm]{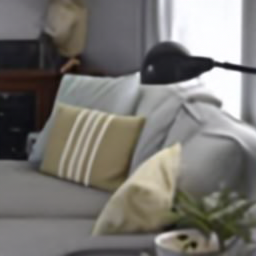}
    \end{subfigure}  
    \caption{
    SRx4 with noise level 0.05. From left to right: original, upsampled observation, DiffPIR (20 NFEs), DDRM (20 NFEs), CM (40 NFEs), CoSIGN (task specific) and our CM4IR (4 NFEs).}
    \label{fig:SRx4_0.1}

\vspace{5mm}

    \centering
    \begin{subfigure}[h]{0.34\columnwidth}
        \centering
        \includegraphics[width=2.8cm, height=2.8cm]{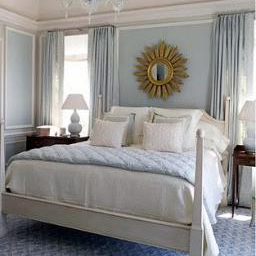}
    \end{subfigure} % 
    \begin{subfigure}[h]{0.34\columnwidth}
        \centering
        \includegraphics[width=2.8cm, height=2.8cm]{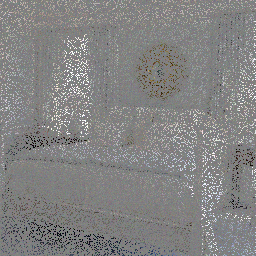}
    \end{subfigure}
    \begin{subfigure}[h]{0.34\columnwidth}
        \centering
        \includegraphics[width=2.8cm, height=2.8cm]{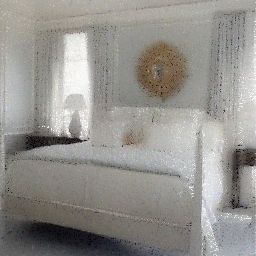}
    \end{subfigure}       
    \begin{subfigure}[h]{0.34\columnwidth}
        \centering
        \includegraphics[width=2.8cm, height=2.8cm]{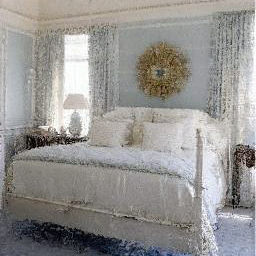}
    \end{subfigure}     
    \begin{subfigure}[h]{0.34\columnwidth}
        \centering
        \includegraphics[width=2.8cm, height=2.8cm]{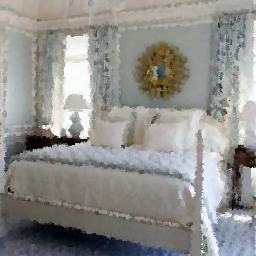}
    \end{subfigure}   
    \begin{subfigure}[h]{0.34\columnwidth} 
        \centering
        \includegraphics[width=2.8cm, height=2.8cm]{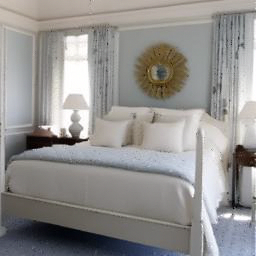}
    \end{subfigure}  
    \caption{
    Inpainting (80\% missing pixels) with noise level 0.05.
    From left to right: original, observation, DiffPIR (20 NFEs), CM (40 NFEs), CoSIGN (task specific) and our CM4IR (4 NFEs).}
    \label{fig:inpainting_0.05}

\vspace{5mm}

    \centering
    \begin{subfigure}[h]{0.34\columnwidth}
        \centering
        \includegraphics[width=2.8cm, height=2.8cm]{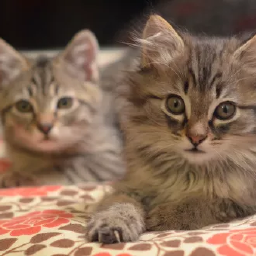}
    \end{subfigure} % 
    \begin{subfigure}[h]{0.34\columnwidth}
        \centering
        \includegraphics[width=2.8cm, height=2.8cm]{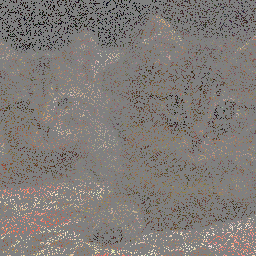}
    \end{subfigure}
    \begin{subfigure}[h]{0.34\columnwidth}
        \centering
        \includegraphics[width=2.8cm, height=2.8cm]{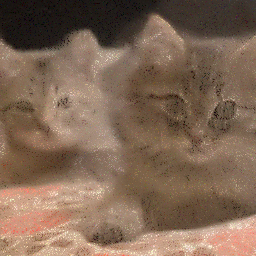}
    \end{subfigure}        
    \begin{subfigure}[h]{0.34\columnwidth}
        \centering
        \includegraphics[width=2.8cm, height=2.8cm]{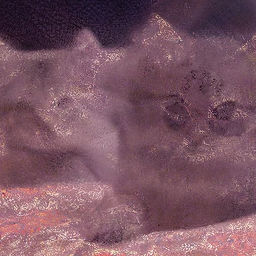}
    \end{subfigure}          
    \begin{subfigure}[h]{0.34\columnwidth}
        \centering
        \includegraphics[width=2.8cm, height=2.8cm]{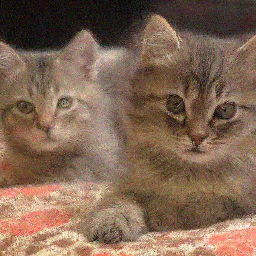}
    \end{subfigure}     
    \begin{subfigure}[h]{0.34\columnwidth}
        \centering
        \includegraphics[width=2.8cm, height=2.8cm]{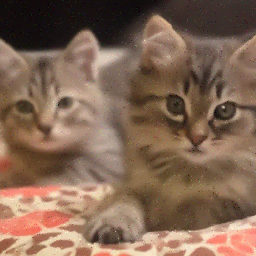}
    \end{subfigure}   
    \caption{
    Inpainting (80\% missing pixels) with noise level 0.05.
    From left to right: original, observation,
    DiffPIR (20 NFEs), DDRM (20 NFEs), CM (40 NFEs) and our CM4IR (4 NFEs).}
    \label{fig:inpainting_0.05_b}
\end{figure*}

\section{Experiments}
\label{sec:exp}

In this section, we examine the performance of CM4IR.
We apply it with $N=4$, so per task there are 4 time points values $\{\tau_n\}_{n=1}^N$ (noise levels) that can be tuned. 
Note that other works that use CMs (for either generation or restoration) \citep{song2023consistency,zhao2024cosign} require careful tuning of the time points.
Yet, to reduce the tuning effort, here we use only two hyperparameters for this goal:  $\gamma>0$ and small $\overline{\alpha}_N>0$, and set $\overline{\alpha}_{n-1}=\overline{\alpha}_n(1+\gamma)$ and $\tau_n = \sqrt{1-\overline{\alpha}_n}$.
As for the other hyperparameters, \tomt{we generally use $\mu_n=1$,  $\eta=0.1$.} % 

\textbf{Datasets.} 
We consider the datasets LSUN bedroom and LSUN cat (256$\times$256), for which there exits both well pretrained CMs \cite{song2023consistency} and DMs (specifically DDPMs) \cite{ho2020denoising}. 
For the test images, we consider 300 validation images for each dataset (as done in \cite{zhao2024cosign}). 
In the appendix 
we also report results for ImageNet 64$\times$64.

\textbf{Tasks.} 
We consider image super-resolution (SR), \tomtb{deblurring} and inpainting tasks, which have been examined also in previous works, \tomtb{and for each of them we consider noise levels of 0.025 and 0.05.}
Specifically, we consider SR with bicubic downsampling of factor 4, 
\tomtb{and deblurring with Gaussian kernel as used, e.g., in \citep{kawar2022denoising,garber2024image}.}
For the inpainting task, we consider the case of random $80\%$ missing pixels % 
and use median init for all the examined methods (we verified that this is beneficial for all of them). 
\tomtb{In the appendix 
we also present results for removal of superimposed text.}  
For each competing method we use hyperparameter settings as suggested by its authors. 
The hyperparameter settings for our CM4IR are stated in the appendix. 
We evaluate the performance of methods by the PSNR and (VGG-)LPIPS \citep{zhang2018unreasonable} metrics.

\subsection{Ablation study}
\label{sec:exp_ablation}

We start by conducting an ablation study of CM4IR on the super-resolution tasks, where we disable parts of Algorithm~\ref{alg:CM_bp}. Specifically, we examine what happens if: % 
1) we do not utilize our noise injection strategy ($\delta=0,\eta=1$); 2) we use $\hat{\z}:=-\hat{\z}^{-}$ (like DDIM \cite{song2020denoising}) instead of $\hat{\z}^{-}$; and 3) we use $\beta\vv$ (i.e., plain Polyak acceleration with tunable $\beta$) instead of $\hat{\z}^{-}$.
{\em Importantly:} in each case we tune the hyperparameters, and in particular the time points (via $\gamma$ and $\overline{\alpha}_N$), separately for best performance.

The PSNR and LPIPS results are presented in Table~\ref{table:results_and_ablation_lsun_bedroom}.
We see the benefit from having the different ingredients in CM4IR. 
In particular, using $\hat{\z}$ instead of $\hat{\z}^{-}$ does not improve much upon the $\eta=1$ case.
Using $\beta\vv$ instead of $\hat{\z}^{-}$ even deteriorates the results compared to the $\eta=1$ case. 
The advantage of using $\hat{\z}^{-}$ over $\hat{\z}$ implies that when having only 4 NFEs, one should not promote closeness of $\x_{\tau_{n-1}}$ to $\x_{\tau_{n}}$. Combining it with $\delta>0$ improves the PSNR \tomt{(higher gains have been observed for inpainting, as shown in the appendix)}.

\subsection{Comparison with other methods}
\label{sec:exp_others}

We turn to compare CM4IR with other methods.
We consider the guided CM scheme proposed in \cite{song2023consistency}, which is based on a predetermined sequence $\{ \tau_n \}$ of size $N=40$ (i.e., 40 NFEs), and thus we name it CM(40). 
The guidance in this scheme resembles BP.  
The major difference of our CM4IR from CM(40) is the noise injection, which allows obtaining good results with a sequence $\{ \tau_n \}$ only of size $N=4$.
We also present the results of CoSIGN \cite{zhao2024cosign}, which is based on supervised fine-tuning to the model per task.
For fair comparison, all the methods are applied with the same CMs. % 

We do not report results for CoSIGN on LSUN cat test set, because its additional ControlNet module is trained only on LSUN bedroom test set per task, and, unsurprisingly, it completely fails (less than 10dB PSNR) on LSUN cat even on the tasks considered in \cite{zhao2024cosign}. 
Another important remark on CoSIGN is that while the tasks in \cite{zhao2024cosign} are stated to include noise level $\sigma_y=0.05$, when inspecting their code we saw that they double the range of the clean signal (to $[-1,1]$) but did not double the noise level as done in the literature \cite{kawar2022denoising,zhu2023denoising,garber2024image}. Therefore, this task-specific method is actually trained for $\sigma_y=0.025$.

We compare our CM4IR also with zero-shot guided DM approaches: DDRM \cite{kawar2022denoising} and DiffPIR \cite{zhu2023denoising}, which require 20 NFEs. % 
We apply these methods with the same DMs 
downloaded from https://huggingface.co/google.

\begin{figure*}
    \center
    \begin{subfigure}[h]{0.34\columnwidth}
        \centering
        \includegraphics[width=2.8cm, height=2.8cm]{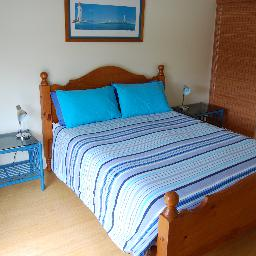}
    \end{subfigure} % 
    \begin{subfigure}[h]{0.34\columnwidth}
        \centering
        \includegraphics[width=2.8cm, height=2.8cm]{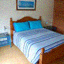}
    \end{subfigure}
    \begin{subfigure}[h]{0.34\columnwidth}
        \centering
        \includegraphics[width=2.8cm, height=2.8cm]{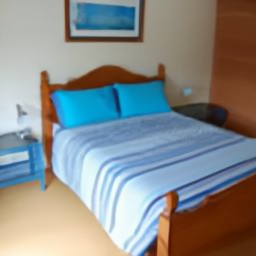}
    \end{subfigure}        
    \begin{subfigure}[h]{0.34\columnwidth}
        \centering
        \includegraphics[width=2.8cm, height=2.8cm]{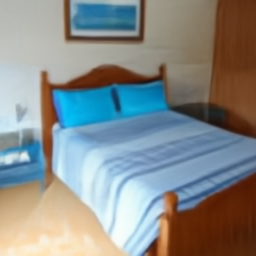}
    \end{subfigure}          
    \begin{subfigure}[h]{0.34\columnwidth}
        \centering
        \includegraphics[width=2.8cm, height=2.8cm]{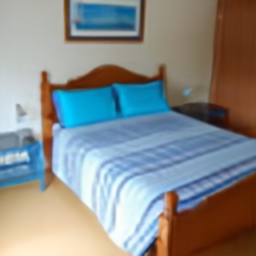}
    \end{subfigure}     
    \begin{subfigure}[h]{0.34\columnwidth}
        \centering
        \includegraphics[width=2.8cm, height=2.8cm]{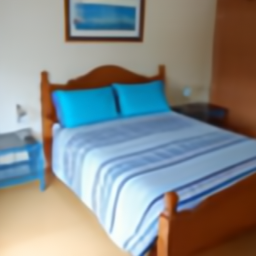}
    \end{subfigure}  

    \caption{
    Super-resolution with noise level 0.025. From left to right: original, observation, DDRM(20 NFEs), DDRM(4 NFEs, auto-calculated), DDRM(4 NFEs, optimized) and DDRM(4 NFEs with our $\hat{\z}^-$ instead of $\hat{\z}$).}
    \label{fig:superresolution_dm_improved}
\end{figure*}

\begin{table*}
\scriptsize % 
\renewcommand{\arraystretch}{1.3}
\caption{Reducing NFEs for DM-based methods.  PSNR [dB] ($\uparrow$) and LPIPS ($\downarrow$) results on LSUN Bedroom validation set.} 
\label{table:reducing_nfes_ddpms}
\centering
    \begin{tabular}{| c | c ||
    c | c | c|c|}
    \hline
 & \diagbox[height=2em,width=10em]{Method}{NFEs, $\{\tau_n\}$} 
 & 20 NFEs 
 & 4 NFEs, auto-calculated
 & 4 NFEs, optimized
 & 4 NFEs with our $\hat{\z}^-$ instead of $\hat{\z}$
 \\ \hline \hline
SRx4, $\sigma_y=0.025$
    & DDRM 
    & 25.67 / {\bf 0.316}
    & 24.16 / 0.395
    & 25.40 / 0.325
    & {\bf 25.89} / 0.327
     \\ \hline

SRx4, $\sigma_y=0.025$
    &DiffPIR 
    & 25.09 / 0.374
    & 24.52 / 0.450
    & 24.68 / 0.425
    & {\bf 25.51} / {\bf 0.371} 
    \\ \hline
    \hline

SRx4, $\sigma_y=0.05$
    & DDRM 
    & 25.08 / {\bf 0.354}
    & 24.12 / 0.396
    & 24.85 / 0.361
    & {\bf 25.22 }/ 0.364
     \\ \hline

SRx4, $\sigma_y=0.05$
    & DiffPIR 
    & 23.83 / 0.457	
    & 23.32 / 0.519
    & 23.42 / 0.506
    & {\bf 24.89} / {\bf 0.404} 
    \\ \hline     

    \end{tabular}
\end{table*}

The results for the LSUN bedrom test set
and the LSUN cat test set are presented in Table \ref{table:results_lsun_bedroom} and
Table \ref{table:results_lsun_cat}, respectively. We present qualitative results % 
in Figures \ref{fig:cat_sr_0.05_intro} - \ref{fig:inpainting_0.05_b}. % 
More results appear in the appendix.

Examining the results, we start with comparing our CM4IR with CM(40) and CoSIGN.
Clearly, CM4IR outperforms CM(40) in all the settings.
Regarding CoSIGN, note that since CoSIGN was specifically trained for the task of SRx4 with bicubic kernel and $\sigma_y=0.025$, we cannot expect to outperform it in this setting (though, CM4IR yields similar PSNR). Yet, the limitations of such a task-specific approach are clearly observed in the other settings: SR with a different noise level, \tomtb{deblurring} and inpainting, % 
\tomtb{where the observation models are not identical to}
those assumed in \cite{zhao2024cosign}.
The lack of flexibility of task-specific approaches is a serious limitation, as in practice, some properties of the degradation model may be unknown apriori or vary at test-time.

\tomtb{Importantly, we see the advantages of our CM4IR over the DM-based zero-shot methods, despite the fact that it requires significantly fewer NFEs. Specifically, CM4IR outperforms DiffPIR in all tasks, and outperforms DDRM in SR and inpainting while being competitive in deblurring.}

\tomtb{Finally, let us state the wall-clock run-time of the zero-shot methods. Using the same hardware -- Nvidia RTX 6000 Ada Generation 48GB -- the average run-time for SRx4 of a single LSUN bedroom image is: 0.159 sec with the proposed CM4IR (4 NFEs), 1.539 sec with CM (40 NFEs), 0.446 sec with DDRM (20 NFEs), 0.659 sec with DiffPIR (20 NFEs). 
Note that: 1) we did not optimize the implementation of our method; (2) DDRM is restricted to tasks where the SVD of $\A$ can be computed and stored efficiently while CM4IR does not require SVD and is applicable in more cases.   
In the appendix, 
we also report performance advantages (better PSNR and LPIPS) of our CM4IR over DPS (1000 NFEs) \cite{chung2022diffusion}, whose per-image run-time on our hardware is 75.461 sec.}

\subsection{Improving few-step guidance of DMs}
\label{sec:exp_DMs}

A key component in our CM4IR is switching the sign of the estimated noise compared to the DDIM method. This hints that applying this modification in guided DM methods that are based on DDIM scheme, such as DDRM \cite{kawar2022denoising} and DiffPIR \cite{zhu2023denoising}, can mitigate their performance drop when drastically reducing the number of the NFEs.

In this section, we explore the performance of DDRM and DiffPIR with 4 NFEs.
We examine three options:
1) we let their official code set the time points (linear spacing in DDRM and quadratic spacing in DiffPIR);
2) we optimize the time points using the procedure described above (via $\gamma$ and $\overline{\alpha}_N$);
and 3) we flip the sign of their injected estimated noise, i.e., use our $\hat{\z}^-$ instead of the DDIM's original $\hat{\z}$ in their implementation and optimize the time points (via $\gamma$ and $\overline{\alpha}_N$).
We state the optimized values in the appendix.

The quantitative results for the SR tasks are presented in Table \ref{table:reducing_nfes_ddpms} \textcolor{black}{and qualitative results are presented in Figure \ref{fig:superresolution_dm_improved}.} 
Noticeable performance drop is observed when reducing the NFE count from 20 to 4 without modification of the noise injection (optimizing the four time points only slightly improves the results). However, utilizing our modified noise injection the performance is boosted, and quite remarkably, the PSNR becomes better than in the original implementation that uses 20 NFEs.
Still, the results are inferior to those of our CM4IR, which further highlights the advantages of using CMs, rather than DMs, for few-step zero-shot restoration.

\section{Conclusion}
\label{sec:conclusion}

In this paper, we proposed a few-step zero-shot restoration scheme, CM4IR, that utilizes the capabilities of Consistency Models (CMs).
We devised our method by combining several ingredients: initialization, back-projection guidance, and above all a novel noise injection mechanism.
Considering image super-resolution, \tomtb{deblurring} and inpainting tasks, we showed that applying our approach with as little as 4 NFEs has already advantages over alternative methods, including zero-shot methods that use many more NFEs of Diffusion Models (DMs).

Interestingly, we also showed that the usefulness of our noise injection technique is not limited to CMs. Specifically, we applied it to improve the performance of existing guided DM methods when drastically reducing NFE number. Yet, these variants did not outperform our CM4IR. 

\section*{Acknowledgment}
\tomtb{The work is supported by ISF (No.~1940/23) and MOST (No.~0007091) grants.}

{
    \small
    \bibliographystyle{ieeenat_fullname}
    \bibliography{refs}
}

\newpage
\onecolumn
\appendix

\section{Proof of Proposition \ref{prop:marginal}}
\label{sec:proofs}

\begin{proposition_non}
For $\eta \in [0,1]$, consider the distribution:
\begin{align}
    q_{\eta}(\x_{\tau_1:\tau_N}|\x_0) = q_{\eta}(\x_{\tau_N}|\x_0) \prod_{n=2}^N q_{\eta}(\x_{\tau_{n-1}}|\x_{\tau_{n}},\x_0)
\end{align}
where $q_{\eta}(\x_{\tau_{N}}|\x_{0}) = \mathcal{N}(\x_0,\tau_{N}^2\I)$, as in \cite{song2020denoising}.
For all $n>1$, let
\begin{align}
\label{eq:ddim_distribution_flip}
    q_{\eta}(\x_{\tau_{n-1}}|\x_{\tau_{n}},\x_{0}) 
    = \mathcal{N}\left( \x_0 + \xi \sqrt{1-\eta^2}\tau_{n-1} \frac{\x_0 - \x_{\tau_n}}{\tau_n} , \eta^2\tau_{n-1}^2\I \right).
\end{align}
where $\xi \in \{-1,1\}$.
Then, for for any $n \in [1,N]$ we have that
\begin{align}
\label{eq:ddim_distribution_marginal}
q_{\eta}(\x_{\tau_{n}}|\x_{0}) = \mathcal{N}(\x_0,\tau_{n}^2\I).
\end{align}
\end{proposition_non}

\begin{proof}

The proof follows the one of Lemma 1 in \cite{song2020denoising}, which is based on induction.
For the base case $n=N$, \eqref{eq:ddim_distribution_marginal} holds by construction, and under the induction assumption for $n$, i.e., $q_{\eta}(\x_{\tau_{n}}|\x_{0}) = \mathcal{N}(\x_0,\tau_{n}^2\I)$, one needs to prove for $n-1$.

As pointed in \cite{song2020denoising}, the marginal $q_{\eta}(\x_{\tau_{n-1}}|\x_{0})=\int_{\x_{\tau_n}}  q_{\eta}(\x_{\tau_{n-1}}|\x_{\tau_{n}},\x_{0}) q_{\eta}(\x_{\tau_{n}}|\x_{0}) d\x_{\tau_n}$ is Gaussian $\mathcal{N}(\bmu,\bSigma)$ with
$$
\bmu = \x_0 + \xi \sqrt{1-\eta^2}\tau_{n-1} \frac{\x_0 - \mathbb{E}\{\x_{\tau_n}|\x_0\}}{\tau_n}
= \x_0 + \xi \sqrt{1-\eta^2}\tau_{n-1} \frac{\x_0 - \x_0}{\tau_n}
= \x_0
$$
$$
\bSigma = \xi^2 \frac{(1-\eta^2)\tau_{n-1}^2}{\tau_n^2} \mathrm{Cov}(\x_{\tau_n}|\x_0) + \eta^2 \tau_{n-1}^2 \I
= \frac{(1-\eta^2)\tau_{n-1}^2}{\tau_n^2} \tau_{n}^2\I + \eta^2 \tau_{n-1}^2 \I = \tau_{n-1}^2\I
$$
where we used $\xi^2=1$.

\end{proof}

\section{More Experimental Details and Results}

In this section, we provide additional details about the experiments, as well as supplementary quantitative and qualitative results that were not included in the main body of the paper due to space constraints.
Our code for reproducing the results will be made available upon acceptance.
Our code is available at
\url{https://github.com/tirer-lab/CM4IR}.

\subsection{Hyperparameter setting} \label{subsec:hyperparameter_settings}

In our experiments, we utilize the DDPM sequence $\{ \beta_i \}_{i=1}^{1000}$ for setting the chosen time points / noise levels. 
Specifically, consistent with many DM methods, the sequence $\{ \beta_i \}$ is defined using a linear schedule that ranges from $\beta_{1}=0.0001$ to $\beta_{{1000}}=0.02$. 
A sequence $\{\hat{\alpha}_i\}$ is computed as follows: $\hat{\alpha}_i = \prod_{j=1}^i \alpha_j$ with $\alpha_j=1-\beta_j$.
Observe that $\{\hat{\alpha}_i\}$ decreases as $i$ increases. 
Given an index number $i_N$, we select the relevant $\hat{\alpha}_{i_N}$ from $\{\hat{\alpha}_i\}$ and set $\overline{\alpha}_N = \hat{\alpha}_{i_N}$.

As explained in Section~\ref{sec:exp}, in CM4IR we use $
\gamma$ and $\overline{\alpha}_N$ to tune the sequence $\{\overline{\alpha}_n\}_{n=1}^N$ for the $N$ NFEs, in the following manner: $\overline{\alpha}_{n-1}=\overline{\alpha}_n(1+\gamma)$.
We keep the sequence bounded within the range $[0, 0.999]$. 

\begin{figure}[htbp]
    \centering
    \begin{subfigure}{0.32\textwidth}
        \centering
        \includegraphics[width=\linewidth]{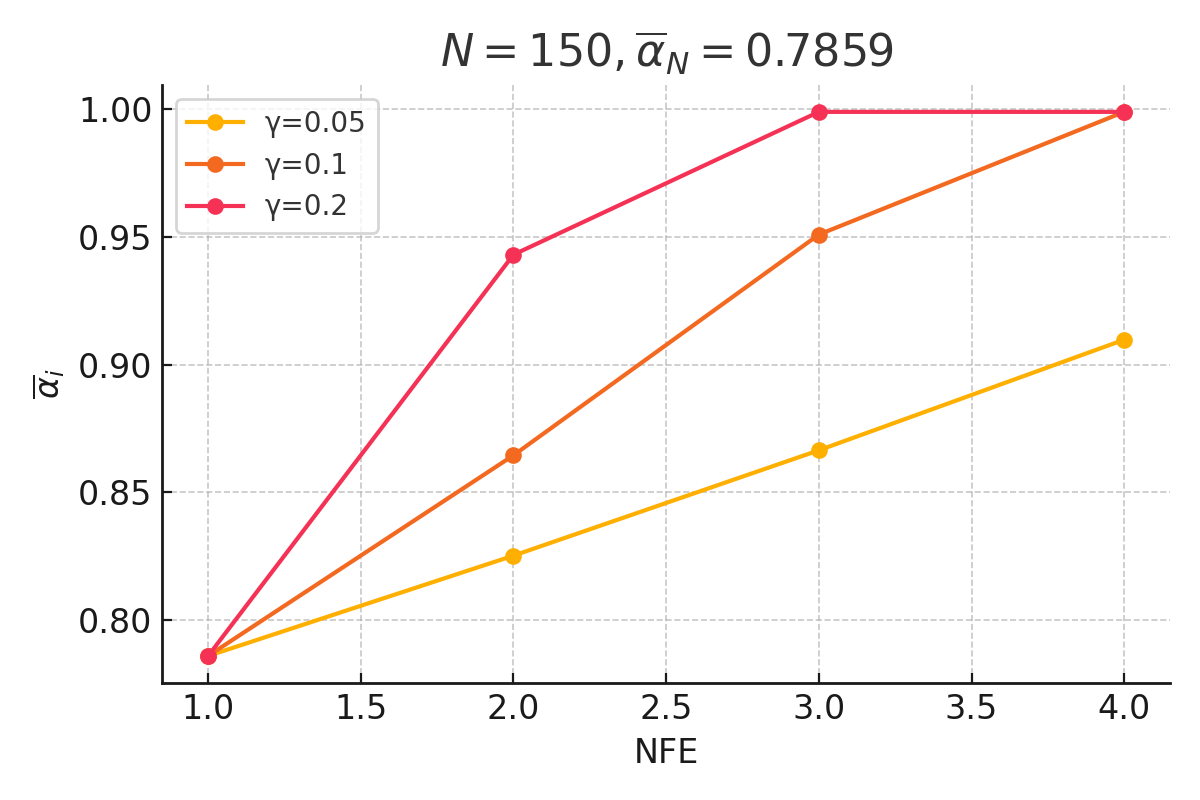}
        \label{fig:alpha_150}
    \end{subfigure}
    \begin{subfigure}{0.32\textwidth}
        \centering
        \includegraphics[width=\linewidth]{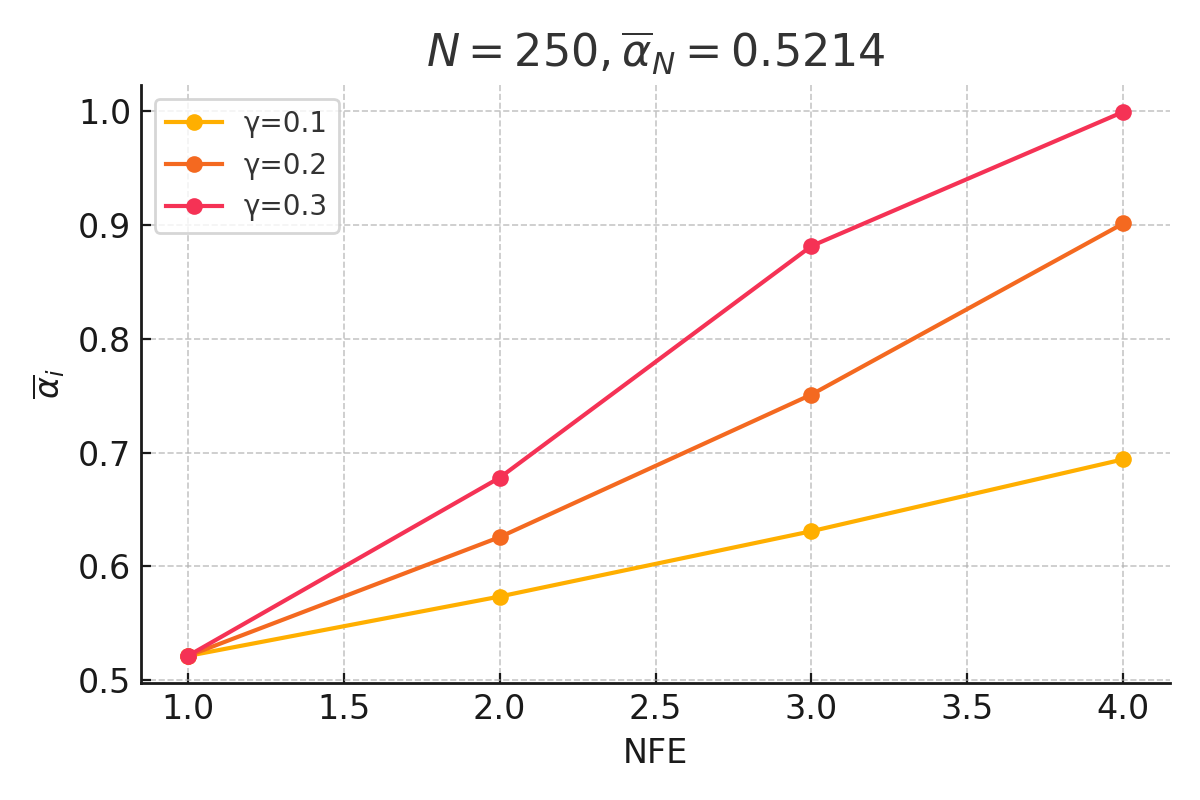}
        \label{fig:alpha_250}
    \end{subfigure}
    \begin{subfigure}{0.32\textwidth}
        \centering
        \includegraphics[width=\linewidth]{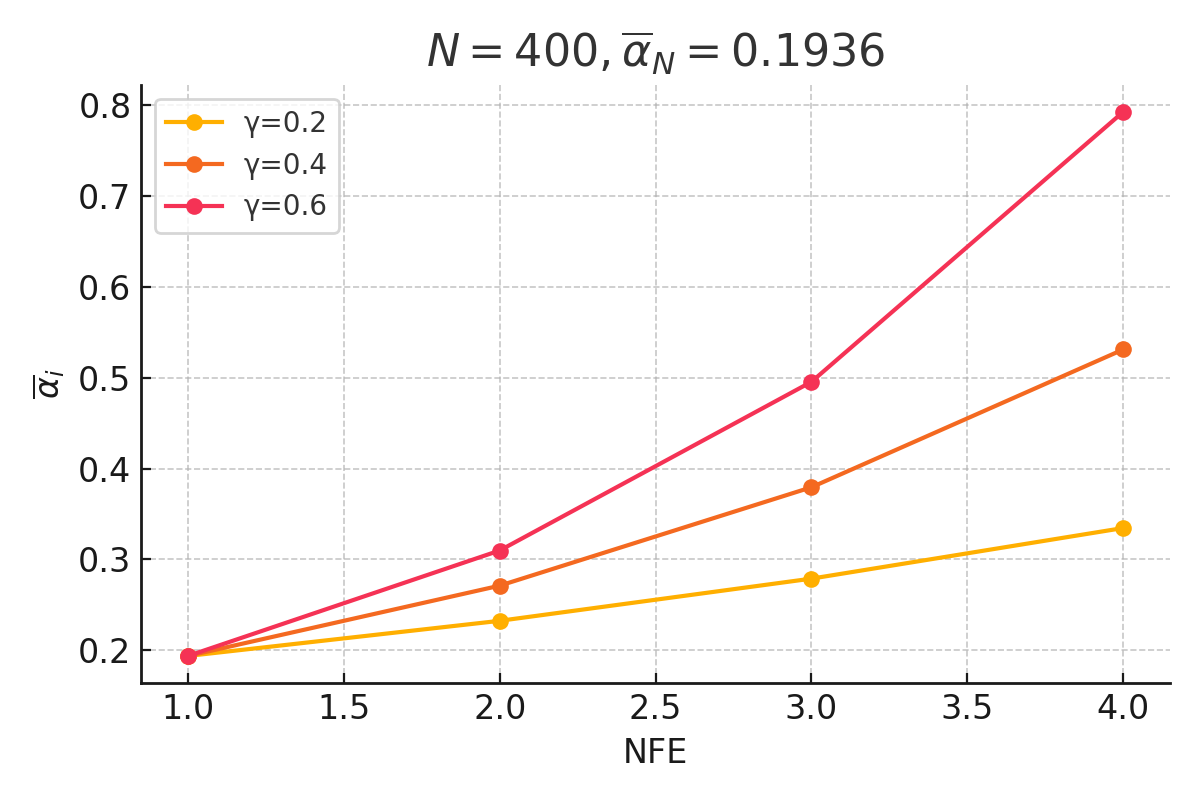}
        \label{fig:alpha_400}
    \end{subfigure}
    \caption{ $\overline{\alpha}_n$ sequences for different $i_N$ and $\gamma$ settings. $\overline{\alpha}_n$ values are clipped to $[0, 0.999]$. Recall that the noise level is $\tau_n=\sqrt{1-\overline{\alpha}_n}$.}
    \label{fig:alpha_sequences}
\end{figure}

In Figure \ref{fig:alpha_sequences}, we observe sequences $\{\overline{\alpha}_n\}$ of length 4 for various $i_N$ values and corresponding $\gamma$ settings. The plots illustrate how the choice of $\gamma$ influences the progression of $\overline{\alpha}_n$ values over time steps, starting from an initial $\overline{\alpha}_N$. 
The left plot corresponds to $i_N=150$ with smaller $\gamma$ values, the middle plot represents $i_N=250$ with moderate $\gamma$ values, and the right plot shows $i_N=400$ with higher $\gamma$ settings. 

The hyperparameters of CM4IR are listed in Table~\ref{table:cm4ir_hyperparams}.
The tuning shows a high level of consistency, making our method intuitive and easy to tune across tasks and datasets. For instance, $\eta$ remains fixed at $0.1$ in all configurations. Similarly, $\gamma$ and $i_N$ values exhibit only minor variations within each task, showing that fine-tuning requires minimal effort. 
\tomtb{For the deblurring task, we also regularize the BP in the intermediate iteration. Namely, we use $\A^T(\A\A^T+\sigma_y^2\zeta)^{-1}$ with $\zeta>0$.} 

\begin{table}[h]
\scriptsize % 
\renewcommand{\arraystretch}{1.3}
\caption{CM4IR hyperparameters.} 
\vspace{-2mm}
\label{table:cm4ir_hyperparams}
\centering
    \begin{tabular}{ | c || c | c |}
    \hline
 Task  & LSUN Bedroom & LSUN Cat \\ \hline \hline

    Bicub.~SRx4~$\sigma_e$=0.025
    & $\eta=0.1$, $i_N=400$, $\gamma=0.7$, $\delta_i=(0.0, 0.5, 0.1, 0.0)$ 
    & $\eta=0.1$, $i_N=400$, $\gamma=0.7$, $\delta_i=(0.0, 0.3, 0.0, 0.0)$  \\ \hline

    Bicub.~SRx4~$\sigma_e$=0.05 
    & $\eta=0.1$, $i_N=250$, $\gamma=0.2$, $\delta_i=(0.0, 0.3, 0.05, 0.1)$ 
    & $\eta=0.1$, $i_N=250$, $\gamma=0.2$, $\delta_i=(0.1, 0.1, 0.0, 0.0)$ 
    \\ \hline
    
    Gauss. Deblurring~$\sigma_e$=0.025 
    & $\eta=0.1$, $i_N=90$, $\gamma=0.02$, $\zeta=3.0$, $\delta_i=(0.0, 0.0, 0.0, 0.0)$ 
     & $\eta=0.1$, $i_N=100$, $\gamma=0.03$,  $\zeta=4.0$, $\delta_i=(0.0, 0.0, 0.0, 0.0)$ 
    \\ \hline
    
    Gauss. Deblurring~$\sigma_e$=0.05 
    & $\eta=0.1$, $i_N=160$, $\gamma=0.07$, $\zeta=1.5$, $\delta_i=(0.0, 0.0, 0.0, 0.0)$ 
     & $\eta=0.1$, $i_N=180$, $\gamma=0.1$,  $\zeta=2.0$, $\delta_i=(0.0, 0.0, 0.0, 0.0)$ 
    \\ \hline
    
    Inpaint. (80\%) ~$\sigma_e$=0
    & $\eta=0.1$, $i_N=150$, $\gamma=0.2$, $\delta_i=(0.1, 0.1, 0.8, 0.8)$ 
    & N/A  \\ \hline
    
    Inpaint. (80\%) ~$\sigma_e$=0.025
    & $\eta=0.1$, $i_N=150$, $\gamma=0.2$, $\delta_i=(0.2, 0.3, 0.8, 0.8)$ 
    & $\eta=0.1$, $i_N=150$, $\gamma=0.2$, $\delta_i=(0.0, 0.0, 1.0, 1.0)$ \\ \hline

    Inpaint. (80\%) ~$\sigma_e$=0.05
    & $\eta=0.1$, $i_N=150$, $\gamma=0.2$, $\delta_i=(0.2, 0.1, 1.0, 1.0)$ 
    & $\eta=0.1$, $i_N=150$, $\gamma=0.2$, $\delta_i=(0.0, 0.0, 1.0, 1.0)$ \\ \hline
    
    \end{tabular}
\end{table}

\subsection{Robustness to hyperparameter settings}
\begin{figure}[ht]
    \centering
    \begin{subfigure}{0.4\textwidth}
        \centering
        \includegraphics[width=\linewidth]{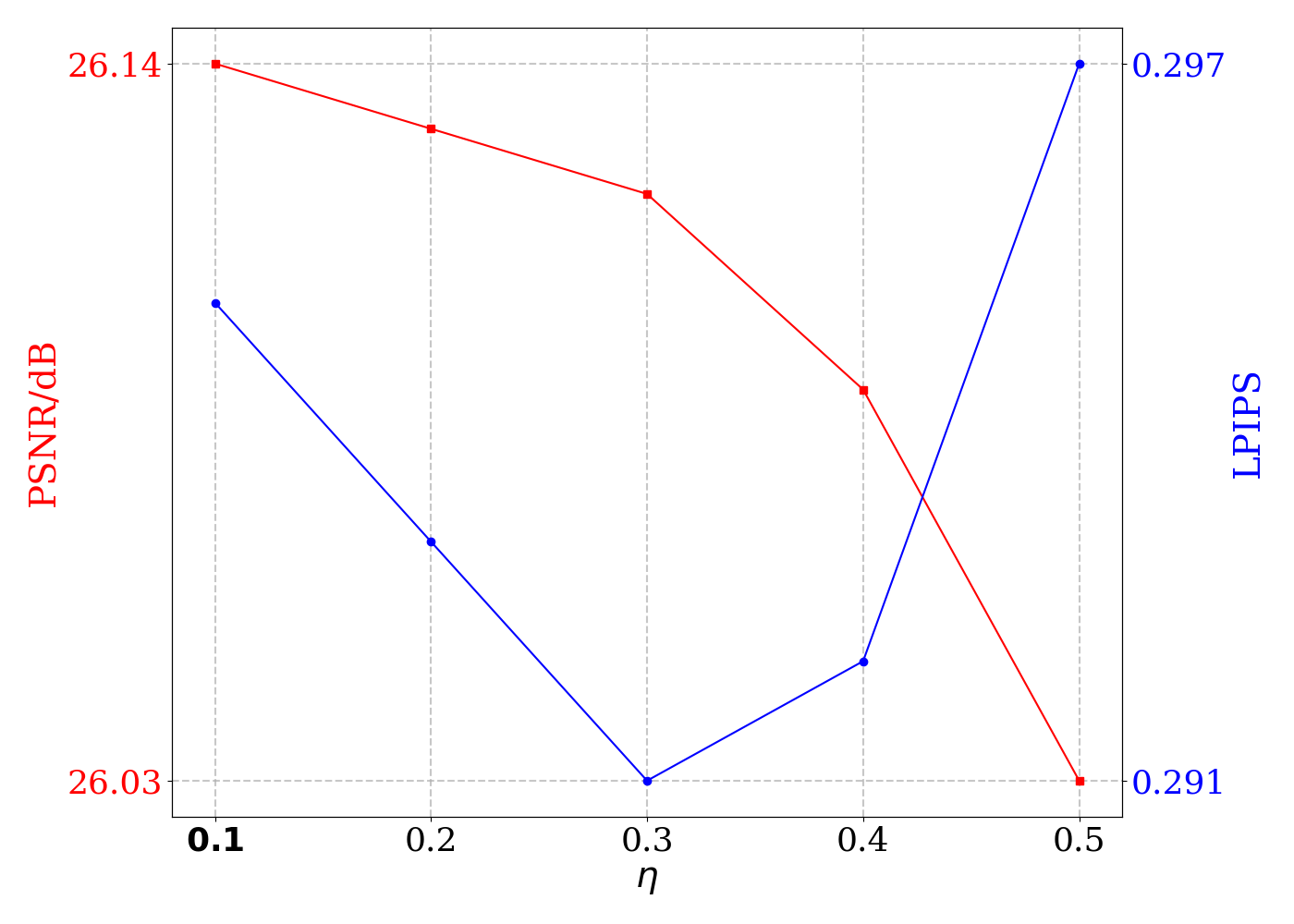}
        \label{fig:eta_vs_psnr_lpips}
    \end{subfigure}
    \begin{subfigure}{0.4\textwidth}
        \centering
        \includegraphics[width=\linewidth]{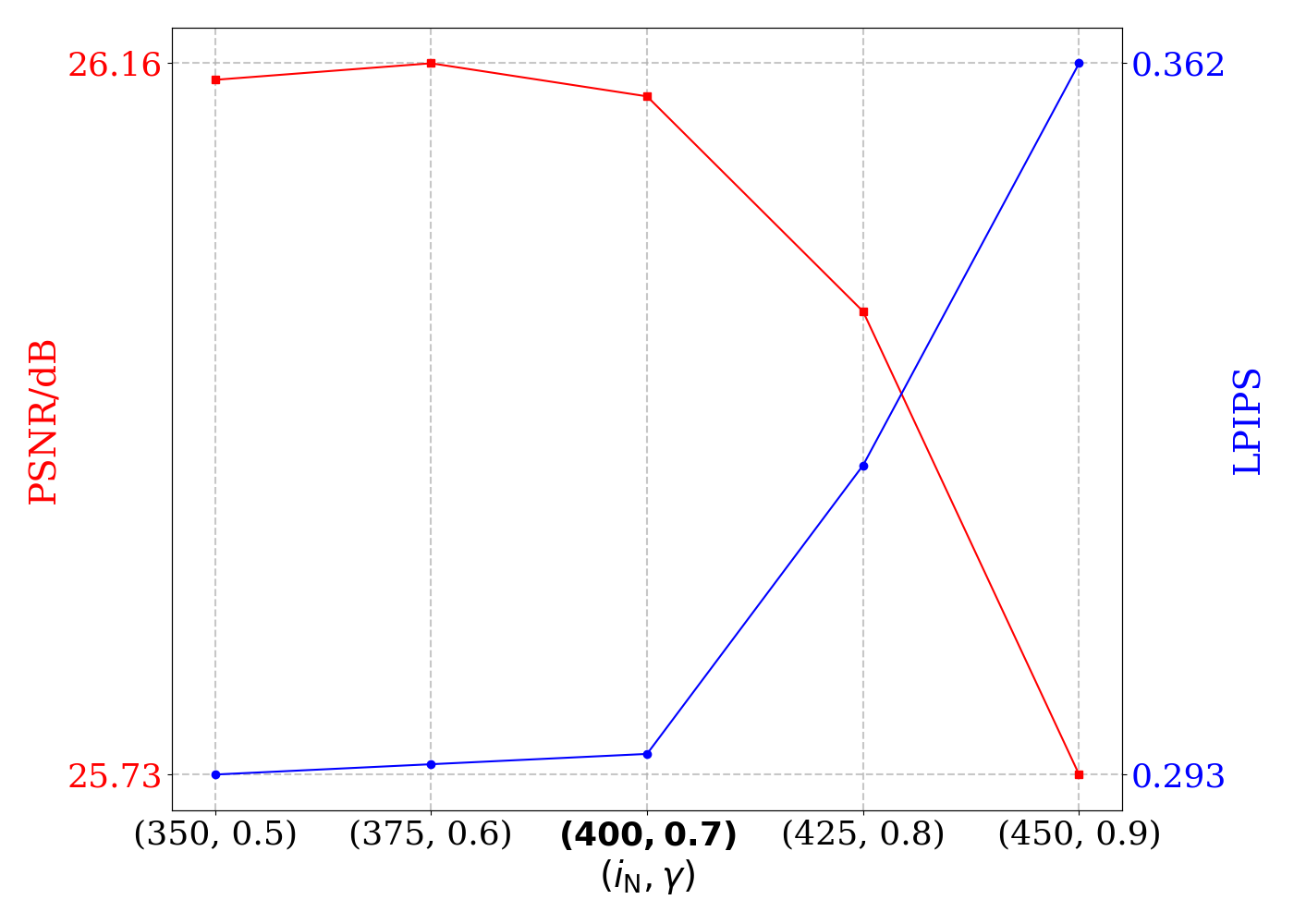}
        \label{fig:gamma_vs_psnr_lpips}
    \end{subfigure}
    
    \caption{ PSNR/LPIPS for different parameters. The chosen values are in \textbf{bold}.}
    \label{fig:params_sensitivity}
\end{figure}

To asses the robustness of CM4IR to the hyperparameter settings, we examine PSNR/LPIPS for SRx4 with $\sigma_y$=0.025 on the LSUN bedroom % 
when we moderately modify the parameters from the values stated in~\ref{subsec:hyperparameter_settings}.  Figure~\ref{fig:params_sensitivity} (left) shows that increasing $\eta$ % 
has only minor effect on the PSNR/LPIPS.
Consistent PSNR/LPIPS are also observed in Figure~\ref{fig:params_sensitivity} (right) when decreasing $(i_N,\gamma)$. % 
Recall that $i_N$ determines the first noise level (higher $i_N$ implies stronger noise) and $\gamma$ determines its decay rate. % 
Thus, it makes sense to % 
pair them,
as examined.

\subsection{Effectiveness of $\delta$}

In Table~\ref{table:results_and_ablation_lsun_bedroom}, we presented some of the effectiveness of using $\delta>0$ within the ablation study.
The results presented in Tables~\ref{table:effectiveness_of_delta_lsun_bedroom} and~\ref{table:effectiveness_of_delta_lsun_cat} demonstrate the effectiveness of using $\delta > 0$ across more tasks and datasets. In most cases, introducing non-zero $\delta$ values consistently improves the PSNR. Additionally, setting $\delta > 0$ often leads to better (or similar) LPIPS.

\begin{table}
\scriptsize % 
\renewcommand{\arraystretch}{1.3}
\caption{Super-resolution, gaussian deblurring and inpainting. PSNR [dB] ($\uparrow$) and LPIPS ($\downarrow$) results on LSUN Bedroom validation set.} 
\label{table:effectiveness_of_delta_lsun_bedroom}
\centering
    \begin{tabular}{ | c || 
    c|c|
    }
    \hline
 \diagbox[height=2em,width=10em]{Task}{$\delta$} 
 & Alg.1 with $\delta=0$ 
 & Alg.1 with $\delta>0$
 \\ \hline \hline

    SRx4~$\sigma_y$=0.025 
    & 25.94 / 0.298
    & {\bf 26.14} / {\bf 0.295}
    \\ \hline

    SRx4~$\sigma_y$=0.05
    &  25.51 /  {\bf 0.320}
    & {\bf 25.60} / {\bf 0.320}
    \\ \hline

    Inpaint. (80\%) $\sigma_y$=0 \
    & 25.13 / 0.296
    & {\bf 25.43} / {\bf 0.284}
    \\ \hline
    
    Inpaint. (80\%) $\sigma_y$=0.025 
    & 25.05 / 0.301 
    & {\bf 25.34} / {\bf 0.295}
    \\ \hline

     Inpaint. (80\%) $\sigma_y$=0.05
    & 24.85 / 0.363 
    & {\bf 25.28} / {\bf 0.328}
    \\ \hline
    \end{tabular}

\vspace{7mm}

\scriptsize % 
\renewcommand{\arraystretch}{1.3}
\caption{Super-resolution and inpainting. PSNR [dB] ($\uparrow$) and LPIPS ($\downarrow$) results on LSUN Cat validation set.} 
\label{table:effectiveness_of_delta_lsun_cat}
\centering
    \begin{tabular}{ | c || 
    c|c|
    }
    \hline
 \diagbox[height=2em,width=10em]{Task}{$\delta$} 
 & Alg.1 with $\delta=0$ 
 & Alg.1 with $\delta>0$ 
 \\ \hline \hline

    SRx4~$\sigma_y$=0.025
    &  27.04 / {\bf 0.325}
    & {\bf 27.18} /  0.328
    \\ \hline
    
    SRx4~$\sigma_y$=0.05 
    & 26.44 / {\bf 0.342}
    & {\bf 26.53} /  0.349
    \\ \hline

    Inpaint. (80\%) $\sigma_y$=0.025
    & 25.75 /  0.381
    & {\bf 25.89} / {\bf 0.364}
    \\ \hline
    
    Inpaint. (80\%) $\sigma_y$=0.05 
    &   25.13 / 0.424
    & {\bf 25.34} / {\bf 0.423}
    \\ \hline
    \end{tabular}
\end{table}

\subsection{Different amount of NFEs}
In this subsection, we investigate alternative choices for the number of NFEs. To provide a comprehensive evaluation, we adapted the same algorithm for setups with 3 NFEs and 5 NFEs, maintaining $\delta=0$ across all cases. Each configuration required tuning of the $i_N$, $\gamma$ and $\eta$ hyperparameters. For $N=3$ NFEs, we tuned $i_N$ and $\gamma$ for obtaining three $\overline{\alpha}_n$ values, and for $N=5$ NFEs we tuned $i_N$ and $\gamma$ for obtaining five $\overline{\alpha}_n$ values.

The results of these experiments are summarized in Table~\ref{table:different_amount_of_nfes_supp}. As the table indicates, while the 3 NFEs configuration achieves reasonable performance, it does not fully utilize the reconstruction capacity of the algorithm. On the other hand, increasing to 5 NFEs does not show gains compared to 4 NFEs. This may be due to restrictions of our tuning strategy that uses only two hyperparameters.

\begin{table}[H]
\scriptsize % 
\renewcommand{\arraystretch}{1.3}
\caption{Evaluating CM4IR with different amount of NFEs. Super-resolution. PSNR [dB] ($\uparrow$) and LPIPS ($\downarrow$) results on LSUN Bedroom validation set.} 
\label{table:different_amount_of_nfes_supp}
\centering
    \begin{tabular}{ | c || c|c|c|c|
    }
    \hline
 \diagbox[height=2em,width=10em]{Task}{NFEs} 
    & 3 NFEs
    & 4 NFEs 
    & 5 NFEs
 \\ \hline \hline
    SRx4~$\sigma_y$=0.025 
    & 25.71 / 0.319 
    & 25.94 / {\bf 0.298}
    & {\bf 25.98} / 0.315
    \\ \hline
    SRx4~$\sigma_y$=0.05 
    & 25.31 . 0.348
    & {\bf 25.51} / {\bf 0.320}
    & 25.42 / 0.339
    \\ \hline

    \end{tabular}
\end{table}

\subsection{Advantages over additional competitors} \label{appendix:dps}

In this subsection, we evaluate DPS \cite{chung2022diffusion}, a reconstruction method based on pretrained DMs with Least-Squares (LS) guidance. For super-resolution, we tested DPS under the settings \( \sigma_y = 0.025 \) and \( \sigma_y = 0.05 \) on the LSUN Bedroom dataset. DPS achieved PSNR/LPIPS results of 24.81 / 0.362 and 24.08 / 0.400 for these noise levels, respectively, which are significantly lower than the performance of our proposed CM4IR method. 

Additionally, DPS requires 1000 NFEs per image, coupled with a much larger per-iteration computational complexity due to computing the model's Jacobian, resulting in an extremely high computational cost. Each image takes approximately 75 seconds to process on the same hardware and using the same DM as the rest of the methods.

In contrast, our CM4IR method operates with only 4 NFEs, reducing the computational time to just 0.159 seconds per image while delivering superior results.

\subsection{Reducing NFEs of additional DM-based method}
DDNM \cite{wang2022zero} is another reconstruction method that leverages pretrained DMs with Back-Projection guidance \cite{tirer2018image}. As highlighted in \cite{garber2024image}, this method is designed and performs well for noiseless observations.
Notice that DDNM utilizes guided DDIM scheme, so like DDRM and DiffPIR, we can explore if its performance drop when reducing the NFEs can be mitigated by our modified noise injection.

In Table~\ref{table:reducing_nfes_sr_noiseless_bedroom}, we present results for reducing the NFEs of DDRM, DiffPIR and DDNM in the noiseless super-resolution task using the LSUN Bedroom validation set. As we can see, our noise injection mechanism improves PSNR, while mitigating LPIPS drops.

\begin{table}[H]
\scriptsize % 
\renewcommand{\arraystretch}{1.3}
\caption{Reducing NFEs for DM-based methods.  PSNR [dB] ($\uparrow$) and LPIPS ($\downarrow$) results on LSUN Bedroom validation set.} 
\label{table:reducing_nfes_sr_noiseless_bedroom}
\centering
    \begin{tabular}{| c | c ||
    c | c | c|c|}
    \hline
 & \diagbox[height=2em,width=10em]{Method}{NFEs, $\{\tau_n\}$} 
 & 20 NFEs 
 & 4 NFEs, auto-calculated
 & 4 NFEs, optimized
 & 4 NFEs with our $\hat{\z}^-$ instead of $\hat{\z}$
 \\ \hline \hline
SRx4, $\sigma_y=0.0$
    & DDRM 
    & 26.27 / {\bf 0.251}
    & 25.90 / 0.272
    & 25.92 / 0.268
    & {\bf 26.31} / 0.261
     \\ \hline

SRx4, $\sigma_y=0.0$
    &DiffPIR 
    & 25.92 / 0.287
    & 25.73 / 0.309
    & 25.69 / 0.319
    & {\bf 26.30} / {\bf 0.271}
    \\ \hline

SRx4, $\sigma_y=0.0$
    & DDNM 
    & 26.33 / {\bf 0.242}
    & 25.93 / 0.268
    & 26.03 / 0.263
    & {\bf 26.38} / 0.262
     \\ \hline

    \end{tabular}
    \vspace{5mm}
\end{table}

\subsection{Additional inpainting results}

In this subsection, we explore additional inpainting scenario - removal of superimposed text.
For reproducibility and fair comparison we use the same text mask, shown in Figure \ref{fig:bedroom_inpaint_letters_0.025_supp}, for all images, and the same median initialization for all methods.
The results in Table \ref{table:results_lsun_bedroom_letters_inpaint_supp} show the advantages of our method.

\begin{table}[h]
\scriptsize % 
\renewcommand{\arraystretch}{1.3}
\caption{Removal of superimposed text. PSNR [dB] ($\uparrow$) and LPIPS ($\downarrow$) results on LSUN Bedroom validation set.} 
\label{table:results_lsun_bedroom_letters_inpaint_supp}
\centering
    \begin{tabular}{ | c || c | c | c | c  | c|c|c|
    c|
    }
    \hline
 \diagbox[height=2em,width=10em]{Task}{Method} 
 & CM (40)  % 
 & CoSIGN (task spec.) 
 & DDRM (20) 
 & DiffPIR (20) 
 & CM4IR (Ours) 
 \\ \hline \hline

    Inpaint. (text)~$\sigma_y$=0.025 
    & 30.86 / 0.189
    & 23.90 / 0.296
    & 33.33 / 0.119
    & 30.95 / 0.198
    & {\bf 34.94} / {\bf 0.087}
    \\ \hline

    \end{tabular}
\end{table}

\begin{figure}
    \centering
    \begin{subfigure}[h]{0.16\textwidth}
        \centering
        \includegraphics[width=2.8cm, height=2.8cm]{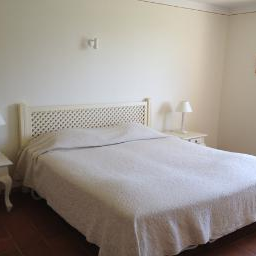}
    \caption*{Original}
    \end{subfigure}
    \begin{subfigure}[h]{0.16\textwidth}
        \centering
        \includegraphics[width=2.8cm, height=2.8cm]{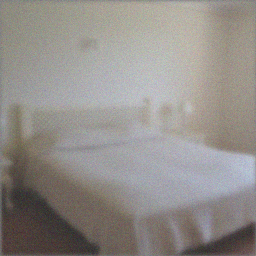}
    \caption*{Observation}
    \end{subfigure}
    \\ % 

    \begin{subfigure}[h]{0.16\textwidth}
        \centering
        \includegraphics[width=2.8cm, height=2.8cm]{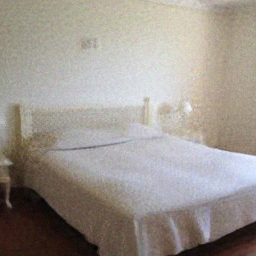}
    \caption*{DiffPIR}
    \end{subfigure}
    \begin{subfigure}[h]{0.16\textwidth}
        \centering
        \includegraphics[width=2.8cm, height=2.8cm]{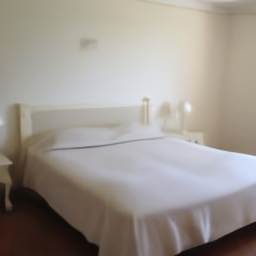}
    \caption*{DDRM}
    \end{subfigure}
    \begin{subfigure}[h]{0.16\textwidth}
        \centering
        \includegraphics[width=2.8cm, height=2.8cm]{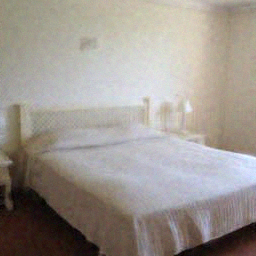}
    \caption*{CM (40)}
    \end{subfigure}
    \begin{subfigure}[h]{0.16\textwidth}
        \centering
        \includegraphics[width=2.8cm, height=2.8cm]{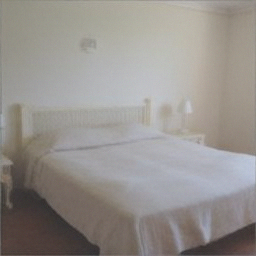}
        \caption*{CoSIGN}
        \end{subfigure}
    \begin{subfigure}[h]{0.16\textwidth}
        \centering
        \includegraphics[width=2.8cm, height=2.8cm]{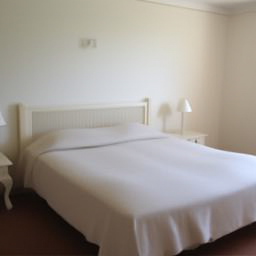}
        \caption*{CM4IR (ours)}
    \end{subfigure}
    \\
    \vspace{5mm}
        \centering
    \begin{subfigure}[h]{0.16\textwidth}
        \centering
        \includegraphics[width=2.8cm, height=2.8cm]{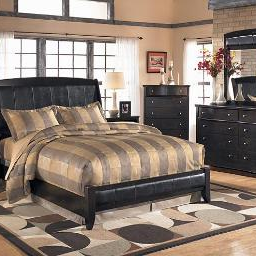}
    \caption*{Original}
    \end{subfigure}
    \begin{subfigure}[h]{0.16\textwidth}
        \centering
        \includegraphics[width=2.8cm, height=2.8cm]{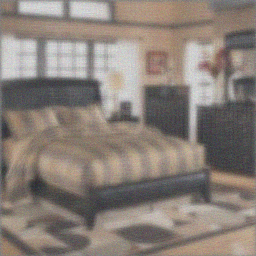}
    \caption*{Observation}
    \end{subfigure}
    \\ % 

    \begin{subfigure}[h]{0.16\textwidth}
        \centering
        \includegraphics[width=2.8cm, height=2.8cm]{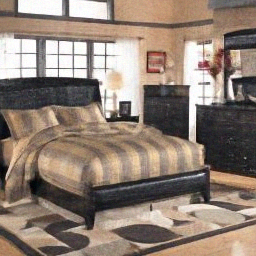}
    \caption*{DiffPIR}
    \end{subfigure}
    \begin{subfigure}[h]{0.16\textwidth}
        \centering
        \includegraphics[width=2.8cm, height=2.8cm]{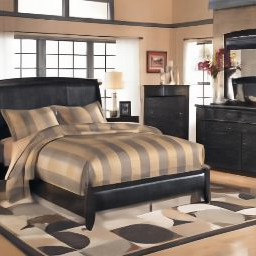}
    \caption*{DDRM}
    \end{subfigure}
    \begin{subfigure}[h]{0.16\textwidth}
        \centering
        \includegraphics[width=2.8cm, height=2.8cm]{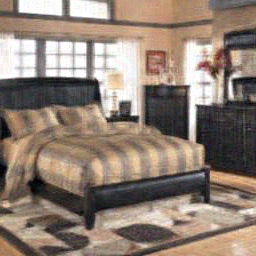}
    \caption*{CM (40)}
    \end{subfigure}
    \begin{subfigure}[h]{0.16\textwidth}
        \centering
        \includegraphics[width=2.8cm, height=2.8cm]{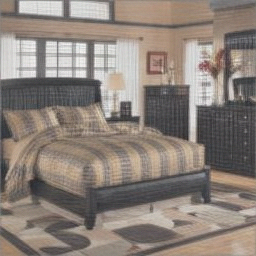}
        \caption*{CoSIGN}
        \end{subfigure}
    \begin{subfigure}[h]{0.16\textwidth}
        \centering
        \includegraphics[width=2.8cm, height=2.8cm]{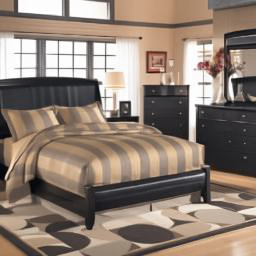}
        \caption*{CM4IR (ours)}
    \end{subfigure}
    \caption{
    Gaussian deblurring with noise level 0.025 }
    \label{fig:bedroom_g_deb_0.025_supp}
\end{figure}

\begin{figure}
    \centering
    \begin{subfigure}[h]{0.16\textwidth}
        \centering
        \includegraphics[width=2.8cm, height=2.8cm]{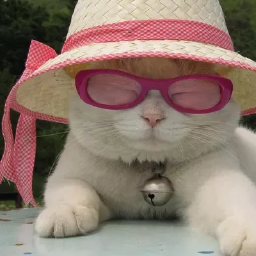}
    \caption*{Original}
    \end{subfigure}
    \begin{subfigure}[h]{0.16\textwidth}
        \centering
        \includegraphics[width=2.8cm, height=2.8cm]{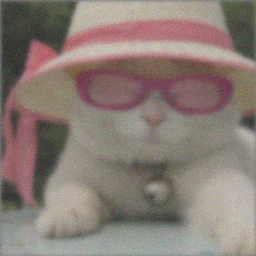}
    \caption*{Observation}
    \end{subfigure}
    \\ % 

    \begin{subfigure}[h]{0.16\textwidth}
        \centering
        \includegraphics[width=2.8cm, height=2.8cm]{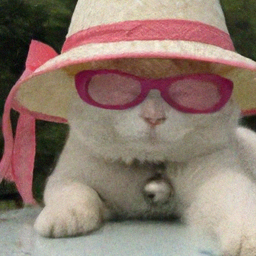}
    \caption*{DiffPIR}
    \end{subfigure}
    \begin{subfigure}[h]{0.16\textwidth}
        \centering
        \includegraphics[width=2.8cm, height=2.8cm]{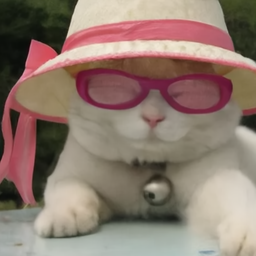}
    \caption*{DDRM}
    \end{subfigure}
    \begin{subfigure}[h]{0.16\textwidth}
        \centering
        \includegraphics[width=2.8cm, height=2.8cm]{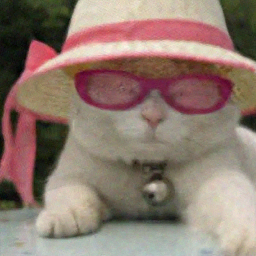}
    \caption*{CM (40)}
    \end{subfigure}
    \begin{subfigure}[h]{0.16\textwidth}
        \centering
        \includegraphics[width=2.8cm, height=2.8cm]{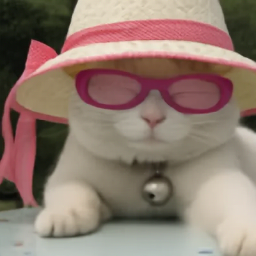}
        \caption*{CM4IR (ours)}
    \end{subfigure}
    \caption{
    Gaussian deblurring with noise level 0.025 }
    \label{fig:cat_g_deb_0.025_supp}
\end{figure}

\begin{figure}
    \centering
    \begin{subfigure}[h]{0.16\textwidth}
        \centering
        \includegraphics[width=2.8cm, height=2.8cm]{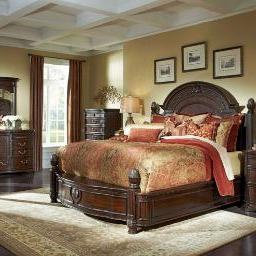}
    \caption*{Original}
    \end{subfigure}
    \begin{subfigure}[h]{0.16\textwidth}
        \centering
        \includegraphics[width=2.8cm, height=2.8cm]{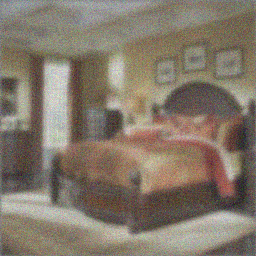}
    \caption*{Observation}
    \end{subfigure}
    \\ % 

    \begin{subfigure}[h]{0.16\textwidth}
        \centering
        \includegraphics[width=2.8cm, height=2.8cm]{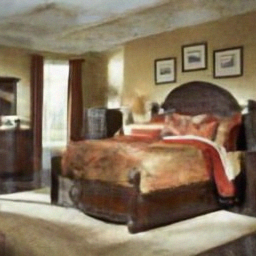}
    \caption*{DiffPIR}
    \end{subfigure}
    \begin{subfigure}[h]{0.16\textwidth}
        \centering
        \includegraphics[width=2.8cm, height=2.8cm]{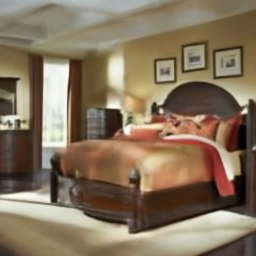}
    \caption*{DDRM}
    \end{subfigure}
    \begin{subfigure}[h]{0.16\textwidth}
        \centering
        \includegraphics[width=2.8cm, height=2.8cm]{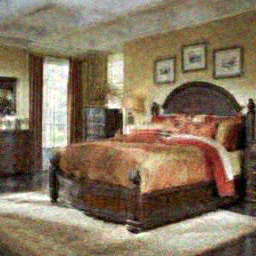}
    \caption*{CM (40)}
    \end{subfigure}
    \begin{subfigure}[h]{0.16\textwidth}
        \centering
        \includegraphics[width=2.8cm, height=2.8cm]{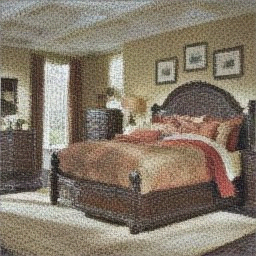}
        \caption*{CoSIGN}
        \end{subfigure}
    \begin{subfigure}[h]{0.16\textwidth}
        \centering
        \includegraphics[width=2.8cm, height=2.8cm]{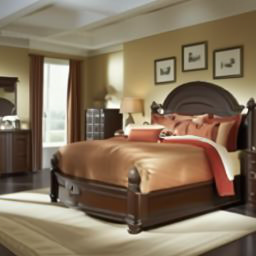}
        \caption*{CM4IR (ours)}
    \end{subfigure}
    \caption{
    Gaussian deblurring with noise level 0.05 }
    \label{fig:bedroom_g_deb_0.05_supp}
\end{figure}

\begin{figure}
    \centering
    \begin{subfigure}[h]{0.16\textwidth}
        \centering
        \includegraphics[width=2.8cm, height=2.8cm]{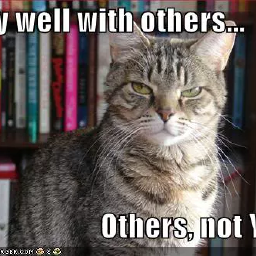}
    \caption*{Original}
    \end{subfigure}
    \begin{subfigure}[h]{0.16\textwidth}
        \centering
        \includegraphics[width=2.8cm, height=2.8cm]{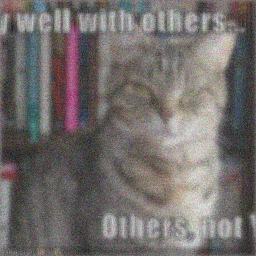}
    \caption*{Observation}
    \end{subfigure}
    \\ % 

    \begin{subfigure}[h]{0.16\textwidth}
        \centering
        \includegraphics[width=2.8cm, height=2.8cm]{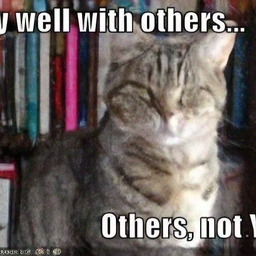}
    \caption*{DiffPIR}
    \end{subfigure}
    \begin{subfigure}[h]{0.16\textwidth}
        \centering
        \includegraphics[width=2.8cm, height=2.8cm]{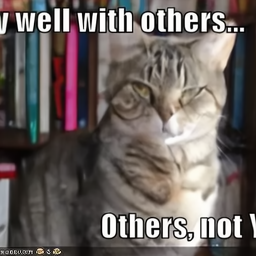}
    \caption*{DDRM}
    \end{subfigure}
    \begin{subfigure}[h]{0.16\textwidth}
        \centering
        \includegraphics[width=2.8cm, height=2.8cm]{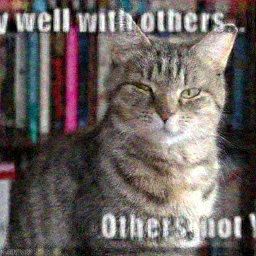}
    \caption*{CM (40)}
    \end{subfigure}
    \begin{subfigure}[h]{0.16\textwidth}
        \centering
        \includegraphics[width=2.8cm, height=2.8cm]{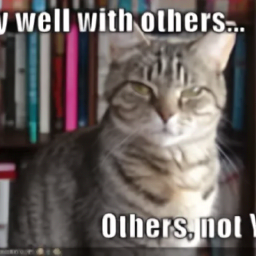}
        \caption*{CM4IR (ours)}
    \end{subfigure}
    \caption{
    Gaussian deblurring with noise level 0.05 }
    \label{fig:cat_g_deb_0.05_supp}
\end{figure}

\begin{figure}
    \centering
    \begin{subfigure}[h]{0.16\textwidth}
        \centering
        \includegraphics[width=2.8cm, height=2.8cm]{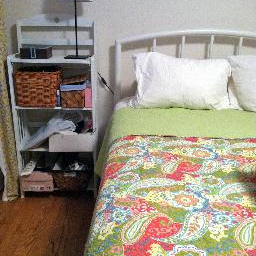}
    \caption*{Original}
    \end{subfigure}
    \begin{subfigure}[h]{0.16\textwidth}
        \centering
        \includegraphics[width=2.8cm, height=2.8cm]{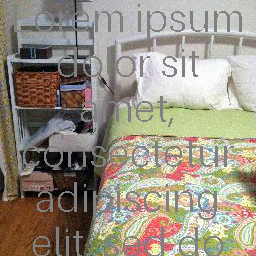}
    \caption*{Observation}
    \end{subfigure}
    \\ % 

    \begin{subfigure}[h]{0.16\textwidth}
        \centering
        \includegraphics[width=2.8cm, height=2.8cm]{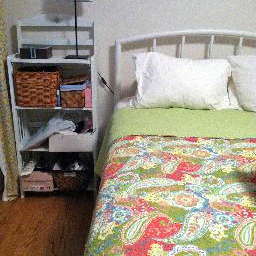}
    \caption*{DiffPIR}
    \end{subfigure}
    \begin{subfigure}[h]{0.16\textwidth}
        \centering
        \includegraphics[width=2.8cm, height=2.8cm]{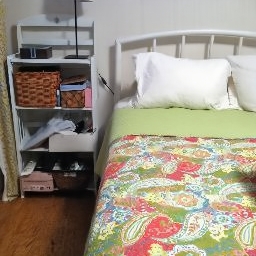}
    \caption*{DDRM}
    \end{subfigure}
    \begin{subfigure}[h]{0.16\textwidth}
        \centering
        \includegraphics[width=2.8cm, height=2.8cm]{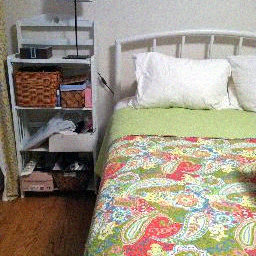}
    \caption*{CM (40)}
    \end{subfigure}
    \begin{subfigure}[h]{0.16\textwidth}
        \centering
        \includegraphics[width=2.8cm, height=2.8cm]{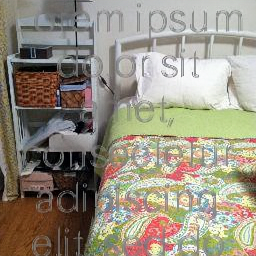}
        \caption*{CoSIGN}
        \end{subfigure}
    \begin{subfigure}[h]{0.16\textwidth}
        \centering
        \includegraphics[width=2.8cm, height=2.8cm]{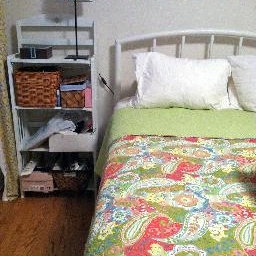}
        \caption*{CM4IR (ours)}
    \end{subfigure}
    \caption{
    Removal of superimposed text with noise level 0.025 }
    \label{fig:bedroom_inpaint_letters_0.025_supp}
\end{figure}

\begin{figure}
    \centering
    \begin{subfigure}[h]{0.16\textwidth}
        \centering
        \includegraphics[width=2.8cm, height=2.8cm]{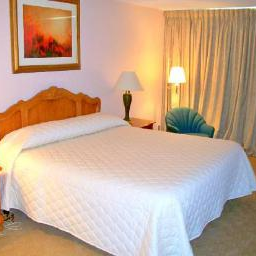}
    \caption*{Original}
    \end{subfigure}
    \begin{subfigure}[h]{0.16\textwidth}
        \centering
        \includegraphics[width=2.8cm, height=2.8cm]{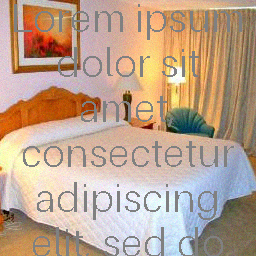}
    \caption*{Observation}
    \end{subfigure}
    \\ % 

    \begin{subfigure}[h]{0.16\textwidth}
        \centering
        \includegraphics[width=2.8cm, height=2.8cm]{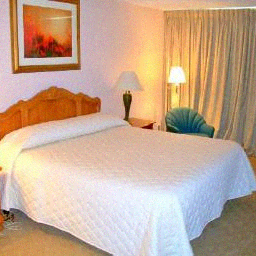}
    \caption*{DiffPIR}
    \end{subfigure}
    \begin{subfigure}[h]{0.16\textwidth}
        \centering
        \includegraphics[width=2.8cm, height=2.8cm]{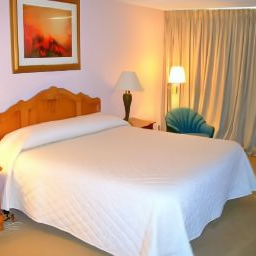}
    \caption*{DDRM}
    \end{subfigure}
    \begin{subfigure}[h]{0.16\textwidth}
        \centering
        \includegraphics[width=2.8cm, height=2.8cm]{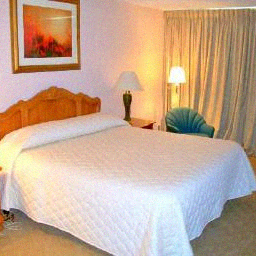}
    \caption*{CM (40)}
    \end{subfigure}
    \begin{subfigure}[h]{0.16\textwidth}
        \centering
        \includegraphics[width=2.8cm, height=2.8cm]{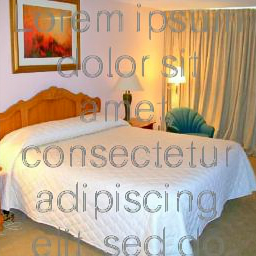}
        \caption*{CoSIGN}
        \end{subfigure}
    \begin{subfigure}[h]{0.16\textwidth}
        \centering
        \includegraphics[width=2.8cm, height=2.8cm]{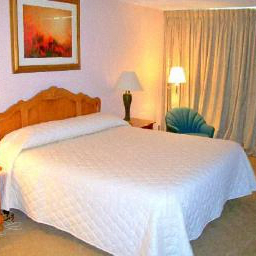}
        \caption*{CM4IR (ours)}
    \end{subfigure}
    \caption{
    Removal of superimposed text with noise level 0.025 }
    \label{fig:bedroom_inpaint_letters_0.025_supp}
\end{figure}

\subsection{Additional dataset}

In this section, we evaluate CM4IR also for ImageNet 64$\times$64 for SR with bicubic downsampling of factor 2 and different noise levels. This factor allows to obtain meaningful recoveries despite the low resolution of the original 64$\times$64 images. The results are presented in Table~\ref{table:imagenet64_sr2}, and show that CM4IR outperforms the other methods in terms of PSNR and has competitive LPIPS despite using only 4 NFEs.

\begin{table}[h]
\scriptsize % 
\renewcommand{\arraystretch}{1.3}
\caption{Bicubic SRx2 for ImageNet 64x64. PSNR [dB] and LPIPS.} 
\label{table:imagenet64_sr2}
\centering
    \begin{tabular}{ | c || c | c | c  | c|c|c|
    c|
    }
    \hline
 \diagbox[height=2em,width=10em]{Task}{Method} 
 & CM (40 NFEs)  
 & DDRM (20 NFEs) % 
 & DiffPIR (20 NFEs) 

 & CM4IR (Ours, 4 NFEs) 
 \\ \hline \hline

    SRx2, $\sigma_y$=0.01 
    &  27.33 / 0.165
    & 29.26 / {\bf 0.104}
    & 28.64 / 0.132
    & {\bf 30.11} / 0.122
    \\ \hline
    
    SRx2, $\sigma_y$=0.025
    &  25.98 / 0.235
    & 28.38 / {\bf 0.134}
    & 27.10 / 0.163
    & {\bf 29.20} / 0.148

    \\ \hline

    SRx2, $\sigma_y$=0.05
    &  24.51 / 0.309
    & 27.02 / {\bf 0.178}
    & 23.65 / 0.250
    & {\bf 27.74} / 0.188
    \\ \hline
    \end{tabular}

\end{table}

\subsection{More qualitative results}
In this subsection, we present visual results for the different tasks.

\begin{figure}
    \centering
    \begin{subfigure}[h]{0.16\textwidth}
        \centering
        \includegraphics[width=2.8cm, height=2.8cm]{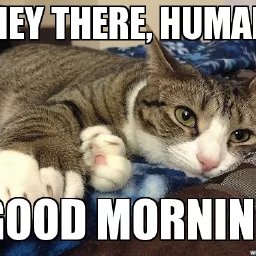}
    \caption*{Original}
    \end{subfigure}
    \begin{subfigure}[h]{0.16\textwidth}
        \centering
        \includegraphics[width=2.8cm, height=2.8cm]{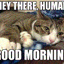}
    \caption*{Observation}
    \end{subfigure}
    \\ % 

    \begin{subfigure}[h]{0.16\textwidth}
        \centering
        \includegraphics[width=2.8cm, height=2.8cm]{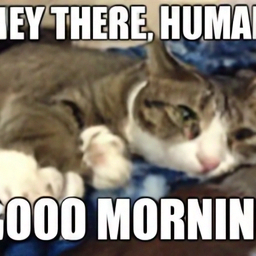}
    \caption*{DiffPIR}
    \end{subfigure}
    \begin{subfigure}[h]{0.16\textwidth}
        \centering
        \includegraphics[width=2.8cm, height=2.8cm]{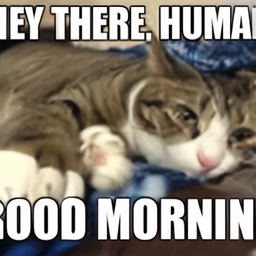}
    \caption*{DDRM}
    \end{subfigure}
    \begin{subfigure}[h]{0.16\textwidth}
        \centering
        \includegraphics[width=2.8cm, height=2.8cm]{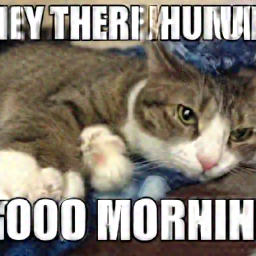}
    \caption*{CM (40)}
    \end{subfigure}
    \begin{subfigure}[h]{0.16\textwidth}
        \centering
        \includegraphics[width=2.8cm, height=2.8cm]{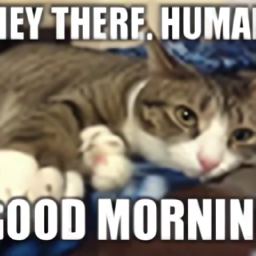}
        \caption*{CM4IR (ours)}
    \end{subfigure}

    \caption{
    Super-resoulution (80\%) with noise level 0.025 }
    \label{fig:cat_sr_0.025_supp}
\end{figure}

\begin{figure}
    \centering
    \begin{subfigure}[h]{0.16\textwidth}
        \centering
        \includegraphics[width=2.8cm, height=2.8cm]{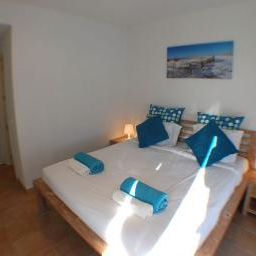}
    \caption*{Original}
    \end{subfigure}
    \begin{subfigure}[h]{0.16\textwidth}
        \centering
        \includegraphics[width=2.8cm, height=2.8cm]{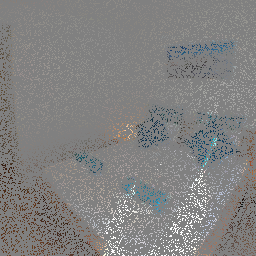}
    \caption*{Observation}
    \end{subfigure}
    \\ % 

    \begin{subfigure}[h]{0.16\textwidth}
        \centering
        \includegraphics[width=2.8cm, height=2.8cm]{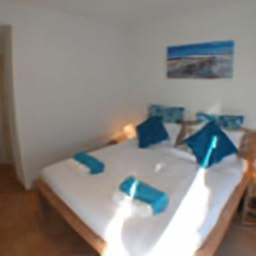}
    \caption*{DiffPIR}
    \end{subfigure}
    \begin{subfigure}[h]{0.16\textwidth}
        \centering
        \includegraphics[width=2.8cm, height=2.8cm]{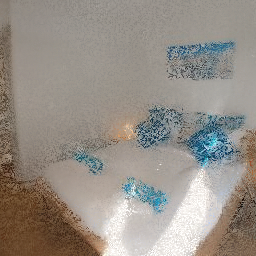}
    \caption*{DDRM}
    \end{subfigure}
    \begin{subfigure}[h]{0.16\textwidth}
        \centering
        \includegraphics[width=2.8cm, height=2.8cm]{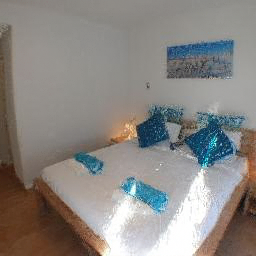}
    \caption*{CM (40)}
    \end{subfigure}
    \begin{subfigure}[h]{0.16\textwidth}
        \centering
        \includegraphics[width=2.8cm, height=2.8cm]{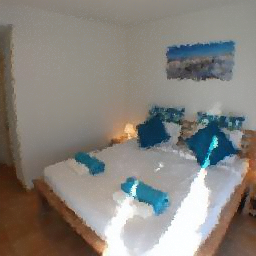}
        \caption*{CoSIGN}
        \end{subfigure}
    \begin{subfigure}[h]{0.16\textwidth}
        \centering
        \includegraphics[width=2.8cm, height=2.8cm]{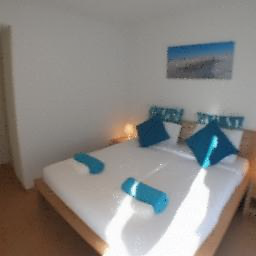}
        \caption*{CM4IR (ours)}
    \end{subfigure}
    \caption{
    Inpainting (80\%) with noise level 0.0 }
    \label{fig:bedroom_inpaint_80_0_supp}
\end{figure}

\begin{figure}
    \centering
    \begin{subfigure}[h]{0.16\textwidth}
        \centering
        \includegraphics[width=2.8cm, height=2.8cm]{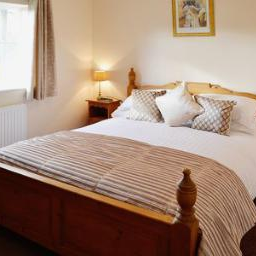}
    \caption*{Original}
    \end{subfigure}
    \begin{subfigure}[h]{0.16\textwidth}
        \centering
        \includegraphics[width=2.8cm, height=2.8cm]{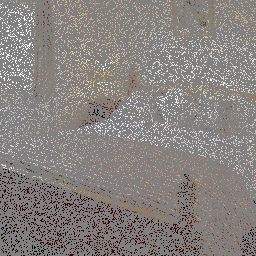}
    \caption*{Observation}
    \end{subfigure}
    \\ % 

    \begin{subfigure}[h]{0.16\textwidth}
        \centering
        \includegraphics[width=2.8cm, height=2.8cm]{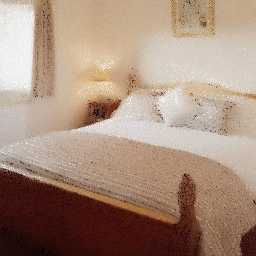}
    \caption*{DiffPIR}
    \end{subfigure}
    \begin{subfigure}[h]{0.16\textwidth}
        \centering
        \includegraphics[width=2.8cm, height=2.8cm]{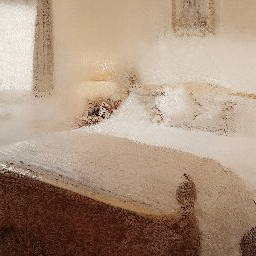}
    \caption*{DDRM}
    \end{subfigure}
    \begin{subfigure}[h]{0.16\textwidth}
        \centering
        \includegraphics[width=2.8cm, height=2.8cm]{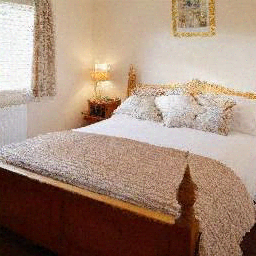}
    \caption*{CM (40)}
    \end{subfigure}
    \begin{subfigure}[h]{0.16\textwidth}
        \centering
        \includegraphics[width=2.8cm, height=2.8cm]{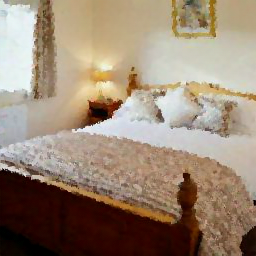}
        \caption*{CoSIGN}
        \end{subfigure}
    \begin{subfigure}[h]{0.16\textwidth}
        \centering
        \includegraphics[width=2.8cm, height=2.8cm]{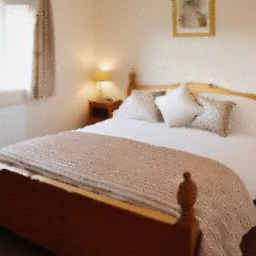}
        \caption*{CM4IR (ours)}
    \end{subfigure}
    \caption{
    Inpainting (80\%) with noise level 0.025 }
    \label{fig:bedroom_inpaint_80_0.025_supp}
\end{figure}

\begin{figure}
    \centering
    \begin{subfigure}[h]{0.16\textwidth}
        \centering
        \includegraphics[width=2.8cm, height=2.8cm]{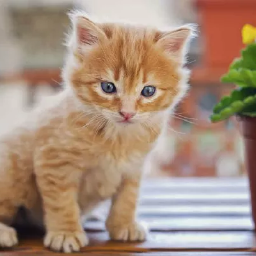}
    \caption*{Original}
    \end{subfigure}
    \begin{subfigure}[h]{0.16\textwidth}
        \centering
        \includegraphics[width=2.8cm, height=2.8cm]{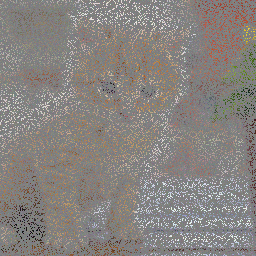}
    \caption*{Observation}
    \end{subfigure}
    \\ % 

    \begin{subfigure}[h]{0.16\textwidth}
        \centering
        \includegraphics[width=2.8cm, height=2.8cm]{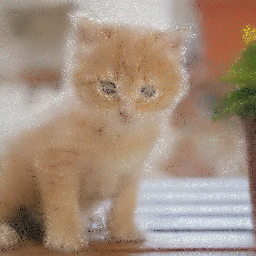}
    \caption*{DiffPIR}
    \end{subfigure}
    \begin{subfigure}[h]{0.16\textwidth}
        \centering
        \includegraphics[width=2.8cm, height=2.8cm]{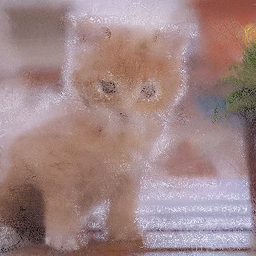}
    \caption*{DDRM}
    \end{subfigure}
    \begin{subfigure}[h]{0.16\textwidth}
        \centering
        \includegraphics[width=2.8cm, height=2.8cm]{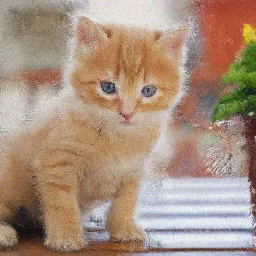}
    \caption*{CM (40)}
    \end{subfigure}
    \begin{subfigure}[h]{0.16\textwidth}
        \centering
        \includegraphics[width=2.8cm, height=2.8cm]{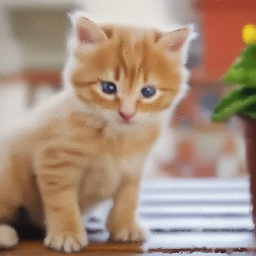}
        \caption*{CM4IR (ours)}
    \end{subfigure}
    \caption{
    Inpainting (80\%) with noise level 0.025 }
    \label{fig:cat_inpaint_80_0.025_supp}
\end{figure}

\begin{figure}
    \centering
    \begin{subfigure}[h]{0.16\textwidth}
        \centering
        \includegraphics[width=2.8cm, height=2.8cm]{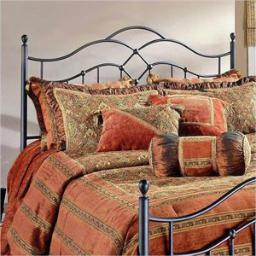}
    \caption{Original}
    \end{subfigure}
    \begin{subfigure}[h]{0.16\textwidth}
        \centering
        \includegraphics[width=2.8cm, height=2.8cm]{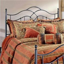}
    \caption{Observation}
    \end{subfigure}
    \begin{subfigure}[h]{0.16\textwidth}
        \centering
        \includegraphics[width=2.8cm, height=2.8cm]{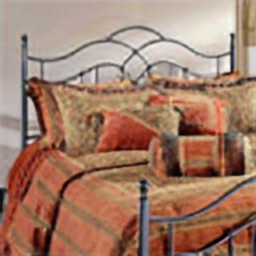}
    \caption{20 NFEs}
    \end{subfigure}
    \begin{subfigure}[h]{0.16\textwidth}
        \centering
        \includegraphics[width=2.8cm, height=2.8cm]{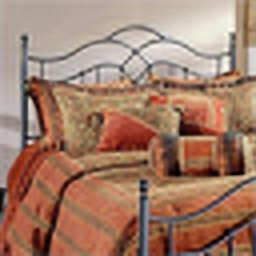}
    \caption{4 NFEs, auto}
    \end{subfigure}
    \begin{subfigure}[h]{0.16\textwidth}
        \centering
        \includegraphics[width=2.8cm, height=2.8cm]{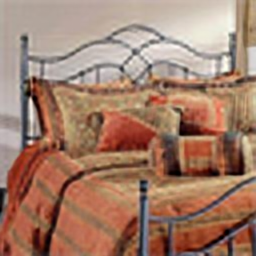}
    \caption{4 NFEs, optimized}
    \end{subfigure}
    \begin{subfigure}[h]{0.16\textwidth}
        \centering
        \includegraphics[width=2.8cm, height=2.8cm]{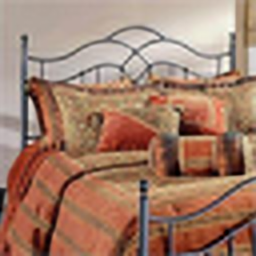}
    \caption{4 NFEs, ours}

    \end{subfigure}

    \caption{
    Reducing NFEs for DDNM, Super-resolution with $\sigma_y=$ 0.0 }
    \label{fig:ddnm_sr_sy_0_reducing_nfes_supp}

\vspace{10mm}

    \centering
    \begin{subfigure}[h]{0.16\textwidth}
        \centering
        \includegraphics[width=2.8cm, height=2.8cm]{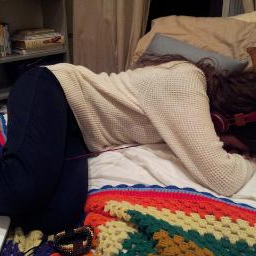}
    \caption{Original}
    \end{subfigure}
    \begin{subfigure}[h]{0.16\textwidth}
        \centering
        \includegraphics[width=2.8cm, height=2.8cm]{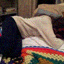}
    \caption{Observation}
    \end{subfigure}
    \begin{subfigure}[h]{0.16\textwidth}
        \centering
        \includegraphics[width=2.8cm, height=2.8cm]{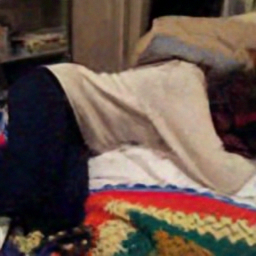}
    \caption{20 NFEs}
    \end{subfigure}
    \begin{subfigure}[h]{0.16\textwidth}
        \centering
        \includegraphics[width=2.8cm, height=2.8cm]{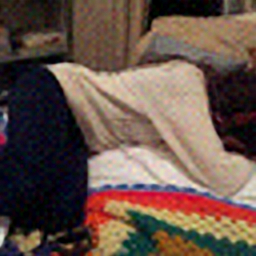}
    \caption{4 NFEs, auto}
    \end{subfigure}
    \begin{subfigure}[h]{0.16\textwidth}
        \centering
        \includegraphics[width=2.8cm, height=2.8cm]{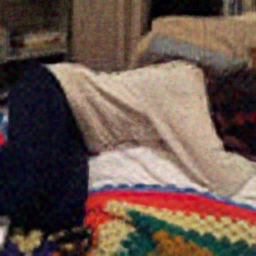}
        \caption{4 NFEs, optimized}
    \end{subfigure}
    \begin{subfigure}[h]{0.16\textwidth}
        \centering
        \includegraphics[width=2.8cm, height=2.8cm]{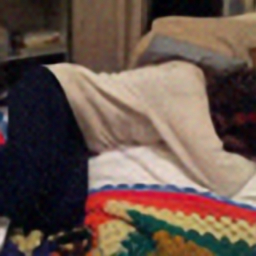}
        \caption{4 NFEs, ours}
    \end{subfigure}
    \caption{
    Reducing NFEs for DiffPIR, Super-resolution with $\sigma_y=$ 0.025 }
    \label{fig:diffpir_sr_sy_0.025_reducing_nfes_supp}

\vspace{10mm}

    \centering
    \begin{subfigure}[h]{0.16\textwidth}
        \centering
        \includegraphics[width=2.8cm, height=2.8cm]{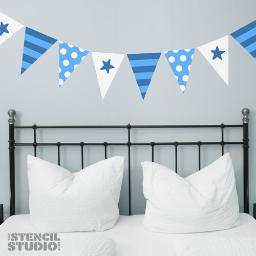}
        \caption{Original}
    \end{subfigure}
    \begin{subfigure}[h]{0.16\textwidth}
        \centering
        \includegraphics[width=2.8cm, height=2.8cm]{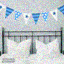}
        \caption{Observation}
    \end{subfigure}
    \begin{subfigure}[h]{0.16\textwidth}
        \centering
        \includegraphics[width=2.8cm, height=2.8cm]{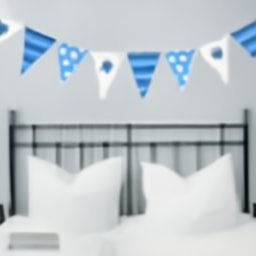}
        \caption{20 NFEs}
    \end{subfigure}
    \begin{subfigure}[h]{0.16\textwidth}
        \centering
        \includegraphics[width=2.8cm, height=2.8cm]{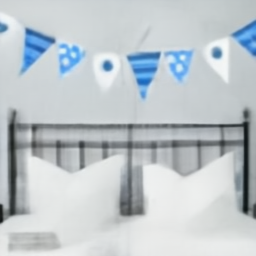}
        \caption{4 NFEs, auto}
    \end{subfigure}
    \begin{subfigure}[h]{0.16\textwidth}
        \centering
        \includegraphics[width=2.8cm, height=2.8cm]{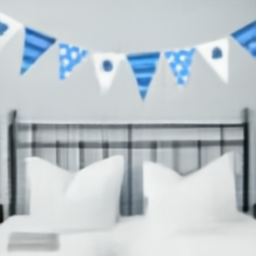}
        \caption{4 NFEs, optimized}
    \end{subfigure}
    \begin{subfigure}[h]{0.16\textwidth}
        \centering
        \includegraphics[width=2.8cm, height=2.8cm]{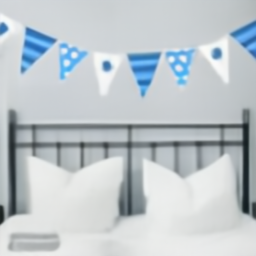}
        \caption{4 NFEs, ours}
    \end{subfigure}  
    \caption{
    Reducing NFEs for DDRM, Super-resolution with $\sigma_y=$ 0.05 }
    \label{fig:ddrm_sr_sy_0.05_reducing_nfes_supp}
\end{figure}

\end{document}